\documentclass[english,superscriptaddress,nofootinbib,citeautoscript]{revtex4-1}
\usepackage[T1]{fontenc}
\usepackage[latin9]{inputenc}
\usepackage{color}
\usepackage{babel}
\usepackage{refstyle}
\usepackage{float}
\usepackage{units}
\usepackage{textcomp}
\usepackage{algorithm2e}
\usepackage{amsmath}
\usepackage{graphicx}
\usepackage[unicode=true,pdfusetitle,
 bookmarks=true,bookmarksnumbered=false,bookmarksopen=false,
 breaklinks=false,pdfborder={0 0 0},pdfborderstyle={},backref=false,colorlinks=true]
 {hyperref}
\hypersetup{
 linkcolor=coolblack,citecolor=burntorange}

\makeatletter

\AtBeginDocument{\providecommand\ref[1]{\ref{app-alg:#1}}}
\newcommand{\lyxdot}{.}

\RS@ifundefined{subsecref}
  {\newref{subsec}{name = \RSsectxt}}
  {}
\RS@ifundefined{thmref}
  {\def\RSthmtxt{theorem~}\newref{thm}{name = \RSthmtxt}}
  {}
\RS@ifundefined{lemref}
  {\def\RSlemtxt{lemma~}\newref{lem}{name = \RSlemtxt}}
  {}

\usepackage{mathrsfs}
\DeclareMathOperator*{\sign}{sign}
\DeclareMathOperator*{\sort}{sort}

\DeclareMathOperator*{\erf}{erf}

\LinesNumbered
\RestyleAlgo{algoruled}
\DontPrintSemicolon

\definecolor{burntorange}{rgb}{0.8, 0.33, 0.0}
\definecolor{charcoal}{rgb}{0.21, 0.27, 0.31}
\definecolor{coolblack}{rgb}{0.0, 0.18, 0.39}

\@ifundefined{showcaptionsetup}{}{%
 \PassOptionsToPackage{caption=false}{subfig}}
\usepackage{subfig}
\makeatother

\begin{document}

\title{Unreasonable Effectiveness of Learning Neural Networks: From Accessible
States and Robust Ensembles to Basic Algorithmic Schemes}

\author{Carlo Baldassi}

\affiliation{Dept. Applied Science and Technology, Politecnico di Torino, Corso
Duca degli Abruzzi 24, I-10129 Torino, Italy}

\affiliation{Human Genetics Foundation-Torino,  Via Nizza 52, I-10126 Torino,
Italy}

\author{Christian Borgs}

\author{Jennifer Chayes}

\affiliation{Microsoft Research, One Memorial Drive, Cambridge, MA 02142, USA}

\author{Alessandro Ingrosso}

\author{Carlo Lucibello}

\author{Luca Saglietti}

\affiliation{Dept. Applied Science and Technology, Politecnico di Torino, Corso
Duca degli Abruzzi 24, I-10129 Torino, Italy}

\affiliation{Human Genetics Foundation-Torino,  Via Nizza 52, I-10126 Torino,
Italy}

\author{Riccardo Zecchina}

\affiliation{Dept. Applied Science and Technology, Politecnico di Torino, Corso
Duca degli Abruzzi 24, I-10129 Torino, Italy}

\affiliation{Human Genetics Foundation-Torino,  Via Nizza 52, I-10126 Torino,
Italy}

\affiliation{Collegio Carlo Alberto, Via Real Collegio 30, I-10024 Moncalieri,
Italy}
\begin{abstract}
In artificial neural networks, learning from data is a computationally
demanding task in which a large number of connection weights are iteratively
tuned through stochastic-gradient-based heuristic processes over a
cost-function. It is not well understood how learning occurs in these
systems, in particular how they avoid getting trapped in configurations
with poor computational performance. Here we study the difficult case
of networks with discrete weights, where the optimization landscape
is very rough even for simple architectures, and provide theoretical
and numerical evidence of the existence of rare\textemdash but extremely
dense and accessible\textemdash regions of configurations in the network
weight space. We define a novel measure, which we call the \emph{robust
ensemble} (RE), which suppresses trapping by isolated configurations
and amplifies the role of these dense regions. We analytically compute
the RE in some exactly solvable models, and also provide a general
algorithmic scheme which is straightforward to implement: define a
cost-function given by a sum of a finite number of replicas of the
original cost-function, with a constraint centering the replicas around
a driving assignment. To illustrate this, we derive several powerful
new algorithms, ranging from Markov Chains to message passing to gradient
descent processes, where the algorithms target the robust dense states,
resulting in substantial improvements in performance. The weak dependence
on the number of precision bits of the weights leads us to conjecture
that very similar reasoning applies to more conventional neural networks.
Analogous algorithmic schemes can also be applied to other optimization
problems.
\end{abstract}
\maketitle
\tableofcontents{}

\section{Introduction}

There is increasing evidence that artificial neural networks perform
exceptionally well in complex recognition tasks \cite{lecun2015deep}.
In spite of huge numbers of parameters and strong non-linearities,
learning often occurs without getting trapped in local minima with
poor prediction performance \cite{ngiam2011optimization}. The remarkable
output of these models has created unprecedented opportunities for
machine learning in a host of applications. However, these practical
successes have been guided by intuition and experiments, while obtaining
a complete theoretical understanding of why these techniques work
seems currently out of reach, due to the inherent complexity of the
problem. In other words: in practical applications, large and complex
architectures are trained on big and rich datasets using an array
of heuristic improvements over basic stochastic gradient descent.
These heuristic enhancements over a stochastic process have the general
purpose of improving the convergence and robustness properties (and
therefore the generalization properties) of the networks, with respect
to what would be achieved with a pure gradient descent on a cost function.

There are many parallels between the studies of algorithmic stochastic
processes and out-of-equilibrium processes in complex systems. Examples
include jamming processes in physics, local search algorithms for
optimization and inference problems in computer science, regulatory
processes in biological and social sciences, and learning processes
in real neural networks (see e.g.~\cite{charbonneau2014fractal,ricci2009cavity,bressloff2014stochastic,easley2010networks,holtmaat2009experience}).
In all these problems, the underlying stochastic dynamics are not
guaranteed to reach states described by an equilibrium probability
measure, as would occur for ergodic statistical physics systems. Sets
of configurations which are quite atypical for certain classes of
algorithmic processes become typical for other processes. While this
fact is unsurprising in a general context, it is unexpected and potentially
quite significant when sets of relevant configurations that are typically
inaccessible for a broad class of search algorithms become extremely
attractive to other algorithms.

Here we discuss how this phenomenon emerges in learning in large-scale
neural networks with low precision synaptic weights. We further show
how it is connected to a novel out-of-equilibrium statistical physics
measure that suppresses the confounding role of exponentially many
deep and isolated configurations (local minima of the error function)
and also amplifies the statistical weight of rare but extremely dense
regions of minima. We call this measure the \emph{Robust Ensemble}
(RE). Moreover, we show that the RE allows us to derive novel and
exceptionally effective algorithms. One of these algorithms is closely
related to a recently proposed stochastic learning protocol used in
complex deep artificial neural networks \cite{NIPS2015_5761}, implying
that the underlying geometrical structure of the RE may provide an
explanation for its effectiveness.

In the present work, we consider discrete neural networks with only
one or two layers, which can be studied analytically. However, we
believe that these results should extend to deep neural networks of
which the models studied here are building blocks, and in fact to
other learning problems as well. We are currently investigating this
\cite{work-in-progress}.

\section{Interacting replicas as a tool for seeking dense regions}

In statistical physics, the canonical ensemble describes the equilibrium
(i.e., long-time limit) properties of a stochastic process in terms
of a probability distribution over the configurations $\sigma$ of
the system $P\left(\sigma;\beta\right)=Z\left(\beta\right)^{-1}\exp\left(-\beta E\left(\sigma\right)\right)$,
where $E\left(\sigma\right)$ is the energy of the configuration,
$\beta$ an inverse temperature, and the normalization factor $Z\left(\beta\right)$
is called the partition function and can be used to derive all thermodynamic
properties. This distribution is thus defined whenever a function
$E\left(\sigma\right)$ is provided, and indeed it can be studied
and provide insight even when the system under consideration is not
a physical system. In particular, it can be used to describe interesting
properties of optimization problems, in which $E\left(\sigma\right)$
has the role of a cost function that one wishes to minimize; in these
cases, one is interested in the limit $\beta\to\infty$, which corresponds
to assigning a uniform weight over the global minima of the energy
function. This kind of description is at the core of the well-known
Simulated Annealing algorithm \cite{kirkpatrick1983optimization}.

In the past few decades, equilibrium statistical physics descriptions
have emerged as fundamental frameworks for studying the properties
of a variety of systems which were previously squarely in the domain
of other disciplines. For example, the study of the phase transitions
of the random $K$-satisfiability problem ($K-$SAT) was linked to
the algorithmic difficulty of finding solutions \cite{mezard2002analytic,krzakala-csp}.
It was shown that the system can exhibit different phases, characterized
by the arrangement of the space of solutions in one, many or a few
connected clusters. Efficient (polynomial-time) algorithms appear
to exist only if the system has so-called ``unfrozen'' clusters:
extensive and connected regions of solutions in which at most a sub-linear
number of variables have a fixed value. If, on the contrary, all solutions
are \textquotedblleft frozen\textquotedblright{} (belonging to clusters
in which an extensive fraction of the variables take a fixed value,
thus confined to sub-spaces of the space of configurations), no efficient
algorithms are known. In the limiting case in which all variables\textemdash except
at most a sub-linear number of them\textemdash are fixed, the clusters
are isolated point-like regions, and the solutions are called \textquotedblleft locked\textquotedblright .\cite{zdeborova2008locked}

\begin{figure}
\includegraphics[width=0.8\columnwidth]{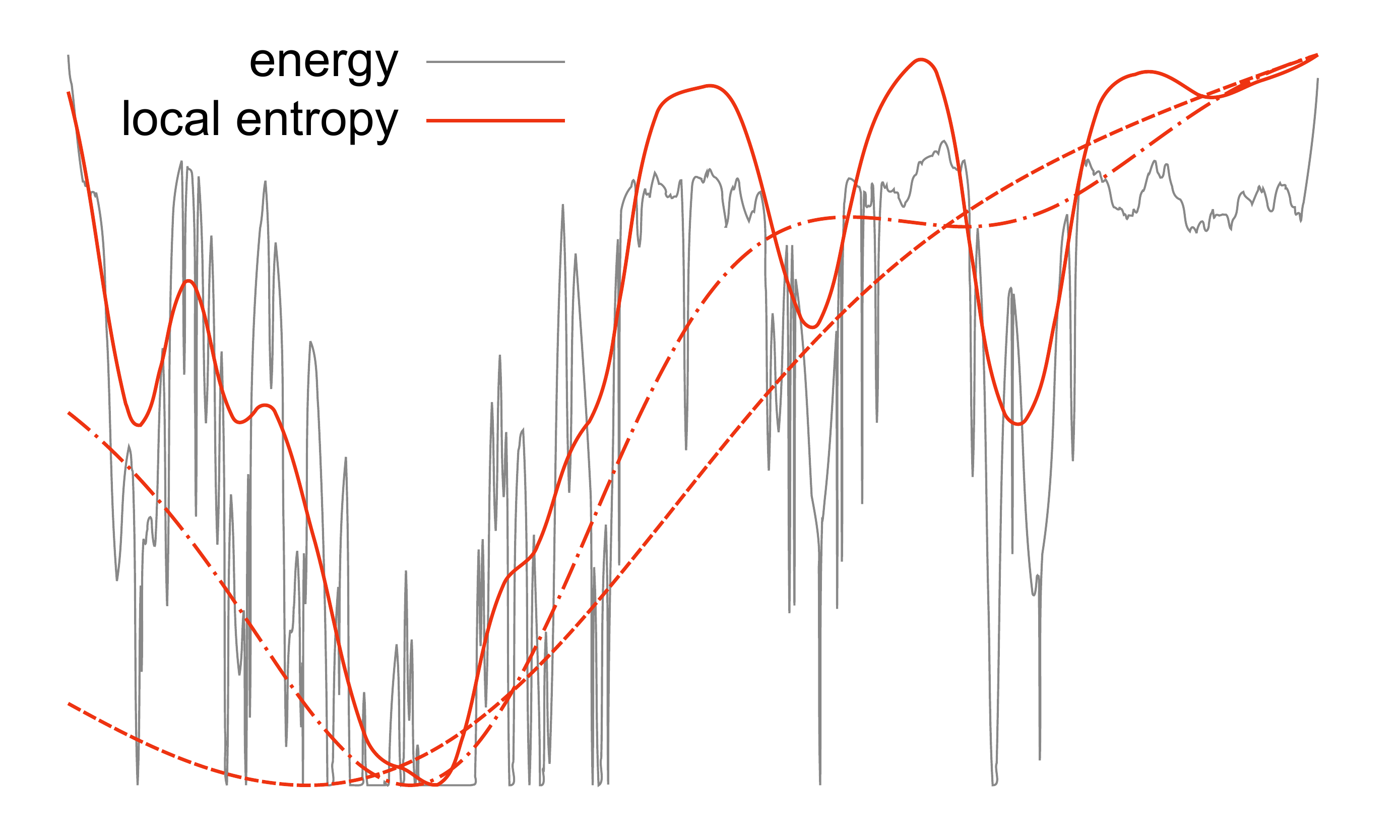}\caption{\label{fig:entropic-landscape-sketch}Energy landscape compared to
local entropy landscape in an illustrative toy example. The energy
landscape (gray curve) can be very rugged, with a large number of
narrow local minima. Some isolated global minima can also be observed
on the right. On the left, there is a region of denser minima which
coalesce into a wide global optimum. The red curves show the local
entropy landscape (eq.~(\ref{eq:local_free_entropy}) with the opposite
sign) computed at increasing values of the interaction parameter $\gamma$,
i.e., at progressively finer scales. At low values of $\gamma$ (dashed
curve), the landscape is extremely smooth and the dense region is
identifiable on a coarse-grained scale. At intermediate values of
$\gamma$ (dot-dashed curve) the global minimum is narrower and located
in a denser region, but it does not correspond to a global energy
minimum yet. At large values of $\gamma$ (solid curve) finer-grain
features appear as several local minima, but the global minimum is
now located inside a wide global optimum of the energy. As a consequence
of this general picture, a process in which a local search algorithm
driven by the local entropic landscape is run at increasing values
of $\gamma$ will end up in such wide minima, even though in the limit
$\gamma\to\infty$ the local entropy landscape tends to the energy
landscape. Note that in a high-dimensional space the isolated global
minima can be exponentially more numerous and thus dominate the equilibrium
measure, but they are ``filtered out'' in the local entropy description.}
\end{figure}

For learning problems with discrete synapses, numerical experiments
indicate that efficient algorithms also seek unfrozen solutions. In
ref.~\cite{baldassi_subdominant_2015}, we showed that the equilibrium
description in these problems is insufficient, in the sense that it
predicts that the problem is always in the completely frozen phase
in which all solutions are locked \cite{huang2014origin}, in spite
of the fact that efficient algorithms seem to exist. This motivated
us to introduce a different measure, which ignores isolated solutions
and enhances the statistical weight of large, accessible regions of
solutions: 
\begin{equation}
P\left(\sigma;\beta,y,\gamma\right)=Z\left(\beta,y,\gamma\right)^{-1}e^{y\,\Phi\left(\sigma,\beta,\gamma\right)}.\label{eq:prob_large_dev}
\end{equation}
Here $y$ is a parameter that has the formal role of an inverse temperature
and $\Phi\left(\sigma,\gamma,\beta\right)$ is a ``local free entropy'':
\begin{equation}
\Phi\left(\sigma,\beta,\gamma\right)=\log\sum_{\left\{ \sigma^{\prime}\right\} }e^{-\beta E\left(\sigma^{\prime}\right)-\gamma\,d\left(\sigma,\sigma^{\prime}\right)}\label{eq:local_free_entropy}
\end{equation}
where $d\left(\cdot,\cdot\right)$ some monotonically increasing function
of the distance between configurations, defined according to the model
under consideration. In the limit $\beta\to\infty$, this expression
reduces (up to an additive constant) to a ``local entropy'': it
counts the number of minima of the energy, weighting them (via the
parameter $\gamma$) by the distance from a reference configuration
$\sigma$. Therefore, if $y$ is large, only the configurations $\sigma$
that are surrounded by an exponential number of local minima will
have a non-negligible weight. By increasing the value of $\gamma$,
it is possible to focus on narrower neighborhoods around $\sigma$,
and at large values of $\gamma$ the reference $\sigma$ will also
with high probability share the properties of the surrounding minima.
This is illustrated in Fig.~\ref{fig:entropic-landscape-sketch}.
These large-deviation statistics seem to capture very well the behavior
of efficient algorithms on discrete neural networks, which invariably
find solutions belonging to high-density regions when these regions
exist, and fail otherwise. These solutions therefore could be \emph{rare}
(i.e., not emerge in a standard equilibrium description), and yet
be \emph{accessible} (i.e., there exist efficient algorithms that
are able to find them), and they are inherently \emph{robust} (they
are immersed in regions of other ``good'' configurations). As discussed
in~\cite{baldassi_subdominant_2015}, there is a relation between
the robustness of solutions in this sense and their good generalization
ability: this is intuitively understood in a Bayesian framework by
considering that a robust solution acts as a representative of a whole
extensive region.

It is therefore natural to consider using our large-deviation statistics
as a starting point to design new algorithms, in much the same way
that Simulated Annealing uses equilibrium statistics. Indeed, this
was shown to work well in ref.~\cite{baldassi_local_2016}. The main
difficulty of that approach was the need to estimate the local (free)
entropy $\Phi$, which was addressed there using the Belief Propagation
(BP) algorithm \cite{mezard_information_2009}.

Here we demonstrate an alternative, general and much simpler approach.
The key observation is that, when $y$ is a non-negative integer,
we can rewrite the partition function of the large deviation distribution
eq.~(\ref{eq:prob_large_dev}) as: 
\begin{eqnarray}
Z\left(\beta,y,\gamma\right) & = & \sum_{\left\{ \sigma^{\star}\right\} }e^{y\,\Phi\left(\sigma^{\star},\beta,\gamma\right)}\label{eq:part_func}\\
 & = & \sum_{\left\{ \sigma^{\star}\right\} }\sum_{\left\{ \sigma^{a}\right\} }e^{-\beta\sum_{a=1}^{y}E\left(\sigma^{a}\right)-\gamma\sum_{a=1}^{y}d\left(\sigma^{\star},\sigma^{a}\right)}\nonumber 
\end{eqnarray}

This partition function describes a system of $y+1$ interacting replicas
of the system, one of which acts as reference while the remaining
$y$ are identical, subject to the energy $E\left(\sigma^{a}\right)$
and the interaction with the reference $\sigma^{\star}$. Studying
the equilibrium statistics of this system and tracing out the replicas
$\sigma^{a}$ is equivalent to studying the original large deviations
model. This provides us with a very simple and general scheme to direct
algorithms to explore robust, accessible regions of the energy landscape:
replicating the model, adding an interaction term with a reference
configuration and running the algorithm over the resulting extended
system.

In fact, in most cases, we can further improve on this scheme by tracing
out the reference instead, which leaves us with a system of $y$ identical
interacting replicas describing what we call the \emph{robust ensemble}
(RE): 
\begin{eqnarray}
Z\left(\beta,y,\gamma\right) & = & \sum_{\left\{ \sigma^{a}\right\} }e^{-\beta\left(\sum_{a=1}^{y}E\left(\sigma^{a}\right)+A\left(\left\{ \sigma^{a}\right\} ,\beta,\gamma\right)\right)}\label{eq:part_func_traced}\\
A\left(\left\{ \sigma^{a}\right\} ,\beta,\gamma\right) & = & -\frac{1}{\beta}\log\sum_{\sigma^{\star}}e^{-\gamma\sum_{a=1}^{y}d\left(\sigma^{\star},\sigma^{a}\right)}\label{eq:interaction_traced}
\end{eqnarray}

In the following, we will demonstrate how this simple procedure can
be applied to a variety of different algorithms: Simulated Annealing,
Stochastic Gradient Descent, and Belief Propagation. In order to demonstrate
the utility of the method, we will focus on the problem of training
neural network models.

\section{Neural network models}

Throughout this paper, we will consider for simplicity one main kind
of neural network model, composed of identical threshold units arranged
in a feed-forward architecture. Each unit has many input channels
and one output, and is parameterized by a vector of ``synaptic weights''
$W$. The output of each unit is given by $\mathrm{sgn}\left(W\cdot\xi\right)$
where $\xi$ is the vector of inputs.

Since we are interested in connecting with analytical results, for
the sake of simplicity all our tests have been performed using binary
weights, $W_{i}^{k}\in\left\{ -1,+1\right\} $, where $k$ denotes
a hidden unit and $i$ an input channel. We should however mention
that all the results generalize to the case of weights with multiple
bits of precision \cite{baldassi2016learning}. We denote by $N$
the total number of synaptic weights in the network, which for simplicity
is assumed to be odd. We studied the random classification problem:
given a set of $\alpha N$ random input patterns $\left\{ \xi^{\mu}\right\} _{\mu=1}^{\alpha N}$,
each of which has a corresponding desired output $\sigma_{D}^{\mu}\in\left\{ -1,+1\right\} $,
we want to find a set of parameters $W$ such that the network output
equals $\sigma_{D}^{\mu}$ for all patterns $\mu$. Thus, for a single-layer
network (also known as a perceptron), the condition could be written
as $\sum_{\mu=1}^{\alpha N}\Theta\left(-\sigma_{D}^{\mu}\left(W\cdot\xi^{\mu}\right)\right)=0$,
where $\Theta\left(x\right)=1$ if $x>0$ and $0$ otherwise. For
a fully-connected two-layer neural network (also known as committee
or consensus machine), the condition could be written as $\sum_{\mu=1}^{\alpha N}\Theta\left(-\sigma_{D}^{\mu}\sum_{k}\mathrm{sgn}\left(W^{k}\cdot\xi^{\mu}\right)\right)=0$
(note that this assumes that all weights in the output unit are $1$,
since they are redundant in the case of binary $W$'s). A three-layer
fully connected network would need to satisfy $\sum_{\mu=1}^{\alpha N}\Theta\left(-\sigma_{D}^{\mu}\sum_{l}\mathrm{sgn}\left(\sum_{k}W_{k}^{2l}\mathrm{sgn}\left(W^{1k}\cdot\xi^{\mu}\right)\right)\right)=0$,
and so on. In this work, we limited our tests to one- and two-layer
networks.

In all tests, we extracted all inputs and outputs in $\left\{ -1,+1\right\} $
from unbiased, identical and independent distributions.

In order to use methods like Simulated Annealing and Gradient Descent,
we need to associate an energy or cost to every configuration of the
system $W$. One natural choice is just to count the number of errors
(mis-classified patterns), but this is not a good choice for local
search algorithms since it hides the information about what direction
to move towards in case of error, except near the threshold. Better
results can be obtained by using the following general definition
instead: we define the energy $E^{\mu}\left(W\right)$ associated
to each pattern $\mu$ as the minimum number of synapses that need
to be switched in order to classify the pattern correctly. The total
energy is then given by the sum of the energy for each pattern, $E\left(W\right)=\sum_{\mu}E^{\mu}\left(W\right)$.
In the single layer case, the energy of a pattern is thus $E^{\mu}\left(W\right)=R\left(-\sigma_{D}^{\mu}\left(W\cdot\xi^{\mu}\right)\right)$,
where $R\left(x\right)=\frac{1}{2}\left(x+1\right)\Theta\left(x\right)$.
Despite the simple definition, the expression for the two-layer case
is more involved and is provided in the Appendix~\ref{app-sub:Energy-definition}.
For deeper networks, the definition is conceptually too naive and
computationally too hard, and it should be amended to reflect the
fact that more than one layer is affected by training, but this is
beyond the scope of the present work.

We also need to define a distance function between replicas of the
system. In all our tests, we used the squared distance $d\left(W,W^{\prime}\right)=\frac{1}{2}\sum_{i=1}^{N}\left(W_{i}-W_{i}^{\prime}\right)^{2}$,
which is proportional to the Hamming distance in the binary case.

\section{Replicated Simulated Annealing}

We claim that there is a general strategy which can be used by a system
of $y$ interacting replicas to seek dense regions of its configuration
space. The simplest example of this is by sampling the configuration
space with a Monte Carlo method \cite{moore2011nature} which uses
the objective functions given by eqs.~(\ref{eq:part_func}) or ~(\ref{eq:part_func_traced}),
and lowering the temperature via a Simulated Annealing (SA) procedure,
until either a zero of the energy (``cost'') or a ``give-up condition''
is reached. For simplicity, we use the RE, in which the reference
configuration is traced out (eq.~(\ref{eq:part_func_traced})), and
we compare our method to the case in which the interaction between
the replicas is absent (i.e.~$\gamma=0$, which is equivalent to
running $y$ parallel independent standard Simulated Annealing algorithms
on the cost function). Besides the annealing procedure, in which the
inverse temperature $\beta$ is gradually increased during the simulation,
we also use a ``scoping'' procedure, which consists in gradually
increasing the interaction $\gamma$, with the effect of reducing
the average distance between the replicas. Intuitively, this corresponds
to exploring the energy landscape on progressively finer scales (Fig.~\ref{fig:entropic-landscape-sketch}).

Additionally, we find that the effect of the interaction among replicas
can be almost entirely accounted for by adding a prior on the choice
of the moves within an otherwise standard Metropolis scheme, while
still maintaining the detailed balance condition (of course, this
reduces to the standard Metropolis rule for $\gamma=0$). The elementary
Monte Carlo iteration step can therefore be summarized as follows:
\emph{i)} we choose a replica uniformly at random; \emph{ii)} we choose
a candidate weight to flip for that replica according to a prior computed
from the state of all the replicas; \emph{iii)} we estimate the single-replica
energy cost of the flip and accept it according to a standard Metropolis
rule. The sampling technique (step \emph{ii} above) and the parameters
used for the simulations are described in the Appendix~\ref{app-sec:RSA}.

\begin{figure}
\includegraphics[width=0.9\textwidth]{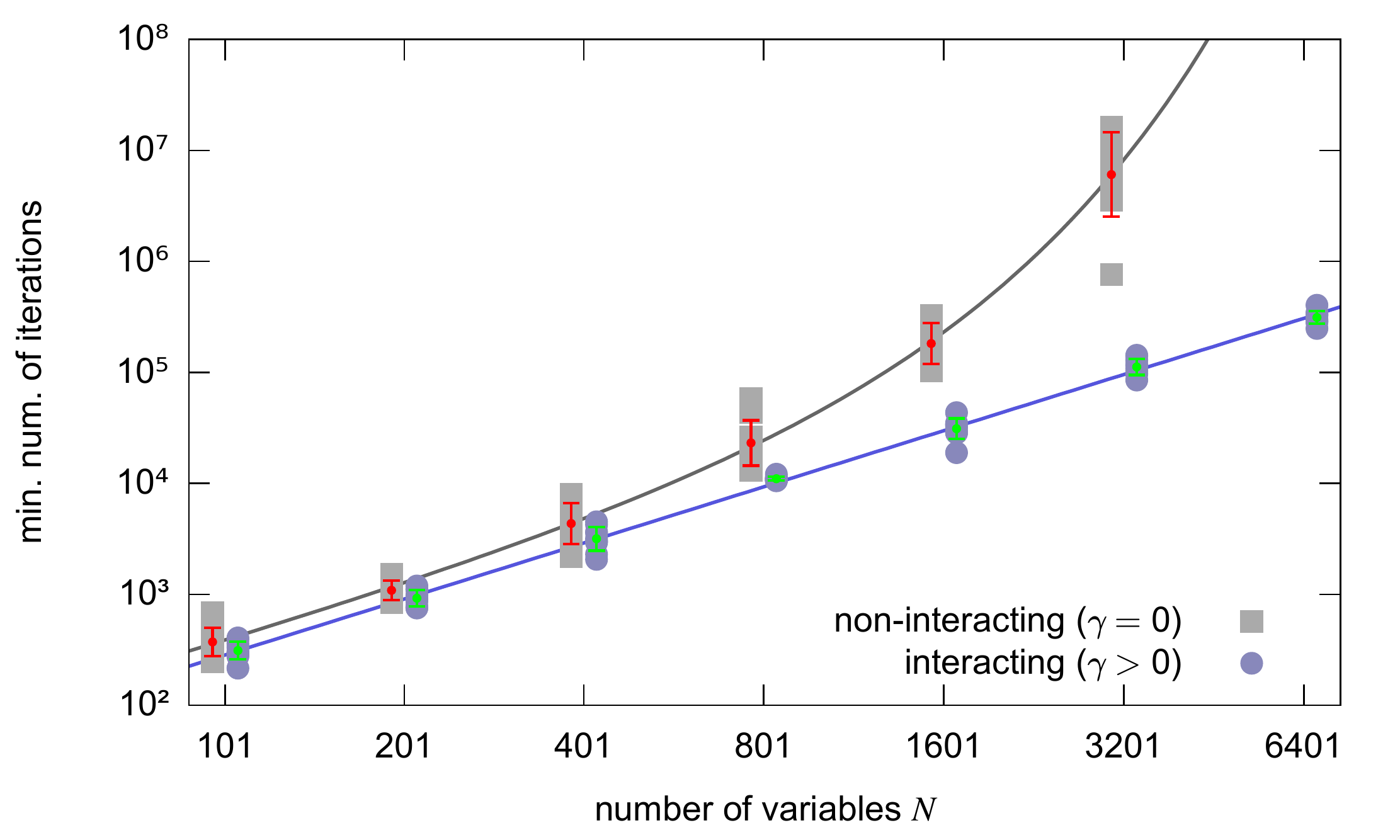}\caption{\label{fig:SimAnnPerc}Replicated Simulated Annealing on the perceptron,
comparison between the interacting version (i.e.~which seeks regions
of high solution density) and the non-interacting version (i.e.~standard
SA), at $\alpha=0.3$ using $y=3$ replicas. With optimized annealing/scoping
parameters, the minimum number of iterations required to find a solution
scales exponentially with $N$ for the standard case, and polynomially
for the interacting case. $10$ samples were tested for each value
of $N$ (the same samples in both cases). The bars represent averages
and standard deviations (taken in logarithmic scale) while the lines
represent fits. The interacting case was fitted by a function $aN^{b}$
with $a\simeq0.13$, $b\simeq1.7$, while the non-interacting case
was fitted by a function $aN^{b}e^{cN^{d}}$ with $a\simeq0.2$, $b\simeq1.5$,
$c\simeq6.6\cdot10^{-4}$ , $d\simeq1.1$. Data is not available for
the non-interacting case at $N=6401$ since we couldn't solve any
of the problems in a reasonable time (the extrapolated value according
to the fit is $\sim3\cdot10^{9}$). The two data sets are slightly
shifted relative to each other for presentation purposes. All the
details are reported in the Appendix~\ref{app-sub:RSA-simulations}.}
\end{figure}

In Fig.~\ref{fig:SimAnnPerc}, we show the results for the perceptron;
an analogous figure for the committee machine, with similar results,
is shown in the Appendix, Fig.~\ref{app-fig:SimAnn}. The analysis
of the scaling with $N$ demonstrates that the interaction is crucial
to finding a solution in polynomial time: the non-interacting version
scales exponentially and it rapidly becomes impossible to find solutions
in reasonable times. Our tests also indicate that the difference in
performance between the interacting and non-interacting cases widens
greatly with increasing the number of patterns per synapse $\alpha$.
As mentioned above, this scheme bears strong similarities to the Entropy-driven
Monte Carlo (EdMC) algorithm that we proposed in~\cite{baldassi_local_2016},
which uses the Belief Propagation algorithm (BP) to estimate the local
entropy around a given configuration. The main advantage of using
a replicated system is that it avoids the need to use BP, which makes
the procedure much simpler and more general. On the other hand, in
systems where BP is able to provide a reasonable estimate of the local
entropy, it can do so directly at a given temperature, and thus avoids
the need to thermalize the replicas. Therefore, the landscapes explored
by the replicated SA and EdMC are in principle different, and it is
possible that the latter has fewer local minima; this however does
not seem to be an issue for the neural network systems considered
here.

\section{Replicated Gradient Descent}

Monte Carlo methods are computationally expensive, and may be infeasible
for large systems. One simple alternative general method for finding
minima of the energy is using Gradient Descent (GD) or one of its
many variants. All these algorithms are generically called backpropagation
algorithms in the neural networks context \cite{rumelhart1988learning}.
Indeed, GD\textemdash in particular Stochastic GD (SGD)\textemdash is
the basis of virtually all recent successful ``deep learning'' techniques
in Machine Learning. The two main issues with using GD are that it
does not in general offer any guarantee to find a global minimum,
and that convergence may be slow (in particular for some of the variables,
cf.~the ``vanishing gradient'' problem \cite{vanishing-gradient}
which affects deep neural network architectures). Additionally, when
training a neural network for the purpose of inferring (generalizing)
a rule from a set of examples, it is in general unclear how the properties
of the local minima of the energy on the training set are related
to the energy of the test set, i.e., to the generalization error.

GD is defined on differentiable systems, and thus it cannot be applied
directly to the case of systems with discrete variables considered
here. One possible work-around is to introduce a two-level scheme,
consisting in using two sets of variables, a continuous one $\mathcal{W}$
and a discrete one $W$, related by a discretization procedure $W=\mathrm{discr}\left(\mathcal{W}\right)$,
and in computing the gradient $\partial E\left(W\right)$ over the
discrete set but adding it to the continuous set: $\mathcal{W}\leftarrow\mathcal{W}-\eta\partial E\left(W\right)$
(where $\eta$ is a gradient step, also called learning rate in the
machine learning context). For the single-layer perceptron with binary
synapses, using the energy definition provided above, in the case
when the gradient is computed one pattern at a time (in machine learning
parlance: using SGD with a minibatch size of $1$), this procedure
leads to the so-called ``Clipped Perceptron'' algorithm (CP). This
algorithm is not able to find a solution to the training problem in
the case of random patterns, but simple (although non-trivial) variants
of it are (SBPI and CP+R, see~\cite{baldassi-et-all-pnas,baldassi-2009}).
In particular CP+R was adapted to two-layer networks (using a a simplified
version of the two-level SGD procedure described above) and was shown
in~\cite{baldassi_subdominant_2015} to be able to achieve near-state-of-the-art
performance on the MNIST database \cite{lecun1998gradient}. The two-level
SGD approach was also more recently applied to multi-layer binary
networks with excellent results in~\cite{courbariaux2015binaryconnect,courbariaux2016binarynet},
along with an array of additional heuristic modifications of the SGD
algorithm that have become standard in application-driven works (e.g.,
batch renormalization). In those cases, however, the back-propagation
of the gradient was performed differently, either because the output
of each unit was not binary \cite{courbariaux2015binaryconnect} or
as a work-around for the use of a different definition for the energy,
which required the introduction of additional heuristic mechanisms
\cite{courbariaux2016binarynet}.

Almost all the above-mentioned results are purely heuristic (except
in the on-line generalization setting, which is not considered in
the present work). Indeed, even just using the two-level SGD is heuristic
in this context. Nevertheless, here we demonstrate that, as in the
case of Simulated Annealing of the previous section, replicating the
system and adding a time-dependent interaction term, i.e., performing
the gradient descent over the robust ensemble (RE) energy defined
in eq.~(\ref{eq:interaction_traced}), leads to a noticeable improvement
in the performance of the algorithm, and that when a solution is found
it is indeed part of a dense region, as expected. We showed in~\cite{baldassi_subdominant_2015}
that solutions belonging to maximally dense regions have better generalization
properties than other solutions; in other words, they are less prone
to overfitting.

It is also important to note here that the stochastic nature of the
SGD algorithm is essential in this case in providing an entropic component
which counters the purely attractive interaction between replicas
introduced by eq.~(\ref{eq:interaction_traced}), since the absolute
minima of the replicated energy of eq.~(\ref{eq:part_func_traced})
are always obtained by collapsing all the replicas in the minima of
the original energy function. Indeed, the existence of an entropic
component allowed us to derive the RE definition by using the interaction
strength $\gamma$ to control the distance via a Legendre transform
in the first place in eq.~(\ref{eq:local_free_entropy}); in order
to use this technique with a non-stochastic minimization algorithm,
the distance should be controlled explicitly instead.

\begin{figure}
\includegraphics[width=0.9\textwidth]{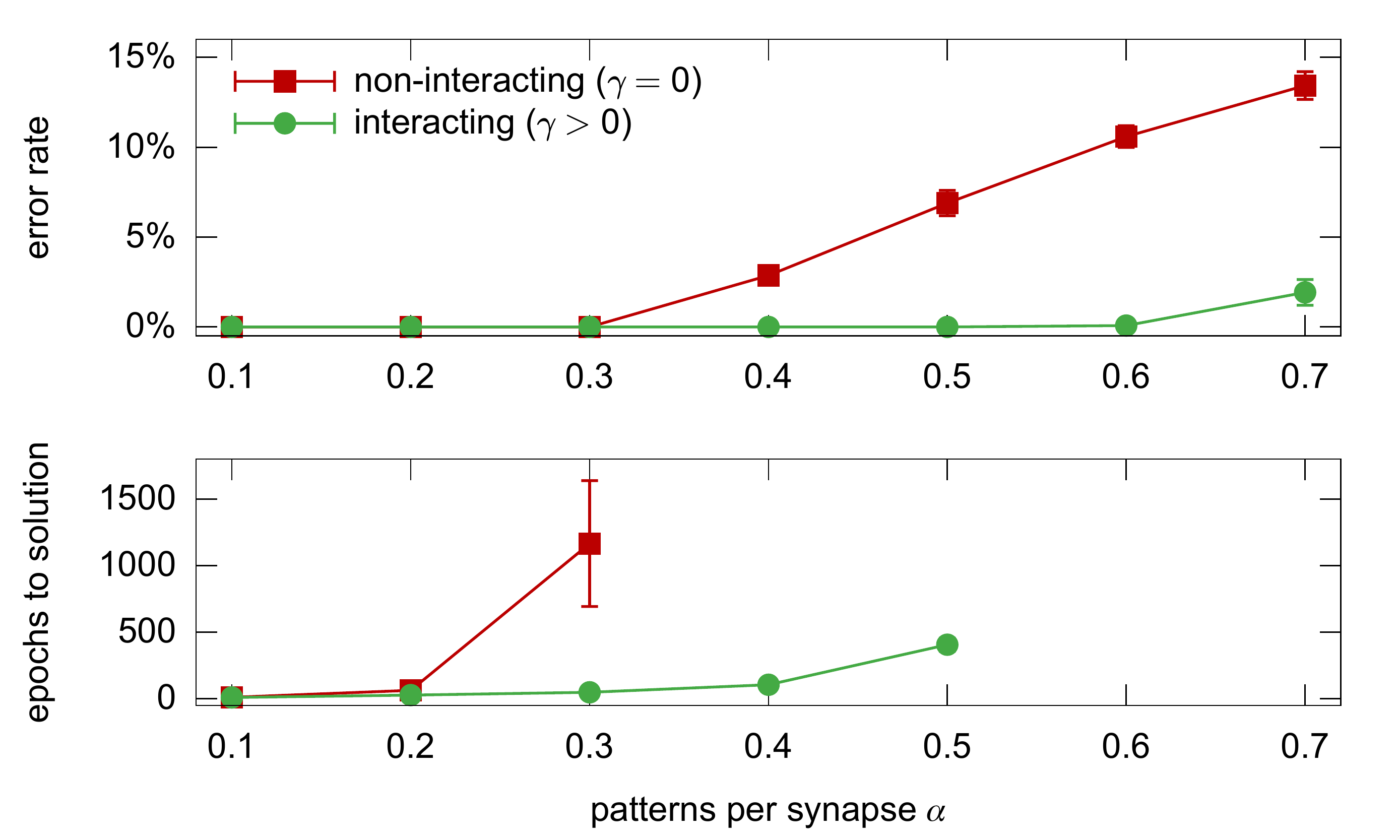}\caption{\label{fig:RSGD_R7_tau80}Replicated Stochastic Gradient descent on
a fully-connected committee machine with $N=1605$ synapses and $K=5$
units in the second layer, comparison between the non-interacting
(i.e.~standard SGD) and interacting versions, using $y=7$ replicas
and a minibatch size of $80$ patterns. Each point shows averages
and standard deviations on $10$ samples with optimal choice of the
parameters, as a function of the training set size. Top: minimum training
error rate achieved after $10^{4}$ epochs. Bottom: number of epochs
required to find a solution. Only the cases with $100\%$ success
rate are shown (note that the interacting case at $\alpha=0.6$ has
$50\%$ success rate but an error rate of just $0.07\%$).}
\end{figure}

In Fig.~\ref{fig:RSGD_R7_tau80} we show the results for a fully
connected committee machine,\footnote{The code used for these tests is available at~\cite{codeRSGD}.}
demonstrating that the introduction of the interaction term greatly
improves the capacity of the network (from $0.3$ to almost $0.6$
patterns per synapse), finds configurations with a lower error rate
even when it fails to solve the problem, and generally requires fewer
presentations of the dataset (epochs). The graphs show the results
for $y=7$ replicas in which the gradient is computed for every $80$
patterns (the so-called minibatch size); we observed the same general
trend for all cases, even with minibatch sizes of $1$ (in the Appendix,
Fig.~\ref{app-fig:RSGD_R3_tau10}, we show the results for $y=3$
and minibatch size $10$). We also observed the same effect in the
perceptron, although with less extensive tests, where this algorithm
has a capacity of at least $0.7$. All technical details are provided
in Appendix~\ref{app-sec:RGD}. These results are in perfect agreement
with the analysis of the next section, on Belief Propagation, which
suggests that this replicated SGD algorithm has near-optimal capacity.

It is interesting to note that a very similar approach\textemdash a
replicated system in which each replica is attracted towards a reference
configuration, called Elastic Averaged SGD (EASGD)\textemdash was
proposed in~\cite{NIPS2015_5761} (see also~\cite{zhang2016distributed})
using deep convolutional networks with continuous variables, as a
heuristic way to exploit parallel computing environments under communication
constraints. Although it is difficult in that case to fully disentangle
the effect of replicating the system from the other heuristics (in
particular the use of ``momentum'' in the GD update), their results
clearly demonstrate a benefit of introducing the replicas in terms
of training error, test error and convergence time. It seems therefore
plausible that, despite the great difference in complexity between
their network and the simple models studied in this paper, the general
underlying reason for the effectiveness of the method is the same,
i.e., the existence of accessible robust low-energy states in the
space of configurations \cite{work-in-progress}.

\section{Replicated Belief Propagation}

Belief Propagation (BP), also known as Sum-Product, is an iterative
message-passing method that can be used to describe a probability
distribution over an instance described by a factor graph in the correlation
decay approximation \cite{mackay2003information,yedidia2005constructing}.
The accuracy of the approximation relies on the assumption that, when
removing an interaction from the network, the nodes involved in that
interaction become effectively independent, an assumption linked to
so-called Replica Symmetry (RS) in statistical physics.

Algorithmically, the method can be briefly summarized as follows.
The goal is to solve a system of equations involving quantities called
messages. These messages represent single-variable \emph{cavity} marginal
probabilities, where the term ``cavity'' refers to the fact that
these probabilities neglect the existence of a node in the graph.
For each edge of the graph there are two messages going in opposite
directions. Each equation in the system gives the expression of one
of the messages as a function of its neighboring messages. The expressions
for these functions are derived from the form of the interactions
between the variables of the graph. The resulting system of equations
is then solved iteratively, by initializing the messages in some arbitrary
configuration and updating them according to the equations until a
fixed point is eventually reached. If convergence is achieved, the
messages can be used to compute various quantities of interest; among
those, the most relevant in our case are the single-site marginals
and the thermodynamic quantities such as the entropy and the free
energy of the system. From single-site marginals, it is also possible
to extract a configuration by simply taking the $\mathrm{argmax}$
of each marginal.

In the case of our learning problem, the variables are the synaptic
weights, and each pattern represents a constraint (i.e., an ``interaction'')
between all variables, and the form of these interactions is such
that BP messages updates can be computed efficiently. Also note that,
since our variables are binary, each of the messages and marginals
can be described with a single quantity: we generally use the magnetization,
i.e., the difference between the probability associated to the states
$+1$ and $-1$. We thus generally speak of the messages as ``cavity
magnetizations'' and the marginals as ``total magnetizations''.
The factor graph, the BP equations and the procedure to perform the
updates efficiently are described in detail in the Appendix section~\ref{app-sub:BP-notes},
closely following the work of~\cite{braunstein-zecchina}. Here,
we give a short summary. We use the notation $m_{j\to\mu}^{t}$ for
the message going from the node representing the weight variable $j$
to the node representing pattern $\mu$ at iteration step $t$, and
$m_{\mu\to j}^{t}$ for the message going in the opposite direction,
related by the BP equation:
\begin{eqnarray}
m_{j\to\mu}^{t} & = & \tanh\sum_{\nu\in\partial j\setminus\mu}\tanh^{-1}m_{\nu\to j}^{t}\label{eq:BP}
\end{eqnarray}
where $\partial j$ indicates the set of all factor nodes connected
to $j$. The expressions for the total magnetizations $m_{j}^{t}$
are identical except that the summation index runs over the whole
set $\partial j$. A configuration of the weights is obtained from
the total magnetizations simply as $W_{j}=\mathrm{sgn}\left(m_{j}\right)$.
The expression for $m_{\nu\to j}^{t+1}$ as a function of $\left\{ m_{i\to\nu}^{t}\right\} _{i=1}^{N}$
is more involved and it's reported in the Appendix section~\ref{app-sub:BP-notes}.

As mentioned above, BP is an inference algorithm: when it converges,
it describes a probability distribution. One particularly effective
scheme to turn BP into a solver is the addition of a ``reinforcement''
term \cite{braunstein-zecchina}: a time-dependent local field is
introduced for each variable, proportional to its own marginal probability
as computed in the previous iteration step, and is gradually increased
until the whole system is completely biased toward one configuration.
This admittedly heuristic scheme is quite general, leads to very good
results in a variety of different problems, and can even be used in
cases in which unmodified BP would not converge or would provide a
very poor approximation (see e.g.~\cite{bailly2011finding}). In
the case of the single layer binary network such as those considered
in this paper, it can reach a capacity of $\alpha\simeq0.75$ patterns
per synapse \cite{braunstein-zecchina}, which is consistent with
the value at which the structure of solution-dense regions breaks
\cite{baldassi_subdominant_2015}.

The reason for the effectiveness of reinforced BP has not been clear.
Intuitively, the process progressively focuses on smaller and smaller
regions of the configuration space, with these regions determined
from the current estimate of the distribution by looking in the ``most
promising'' direction. This process has thus some qualitative similarities
with the search for dense regions described in the previous sections.
This analogy can be made precise by writing the BP equations for the
system described by eq.~(\ref{eq:part_func}). There are in this
case two equivalent approaches: the first is to use the local entropy
as the energy function, using a second-level BP to estimate the local
entropy itself. This approach is very similar to the so called 1-step
replica-symmetry-breaking (1RSB) cavity equations (see ~\cite{mezard_information_2009}
for a general introduction). The second approach is to replicate the
system, considering $N$ vector variables $\left\{ W_{j}^{a}\right\} _{a=1}^{y}$
of length $y$, and assuming an internal symmetry for each variable,
i.e.~that all marginals are invariant under permutation of the replica
indices, similarly to what is done in~\cite{kabashima2005}: $P_{j}\left(\left\{ W_{j}^{a}\right\} _{a=1}^{y}\right)=P_{j}\left(\sum_{a=1}^{y}W_{j}^{a}\right)$.
The result in the two cases is the same (this will be shown in a technical
work, in preparation, where the connection between the large deviations
measure and the 1RSB equilibrium description is also made explicit).
Since BP assumes replica symmetry, the resulting message passing algorithm
reproduces quite accurately the analytical results at the RS level.
As explained in~\cite{baldassi_subdominant_2015}, these results
can however become wrong, in particular for high values of $\alpha$,
$\gamma$ and $y$, due to the onset of correlations (the so called
replica-symmetry-breaking \textendash{} RSB \textendash{} effect \cite{mezard_information_2009}).
More specifically, in this model the RS solution assumes that there
is a single dense region comprising the robust ensemble (RE), while
the occurrence of RSB implies that there are several maximally dense
regions. As a consequence this algorithm is not a very good candidate
as a solver. A more correct description\textemdash which could then
lead to a more controlled solver\textemdash would thus require a third
level of BP equations, or equivalently an assumption of symmetry-breaking
in the structure of the marginals $P_{j}\left(\left\{ W_{j}^{a}\right\} _{a=1}^{y}\right)$.

\begin{figure}
\includegraphics[width=0.9\textwidth]{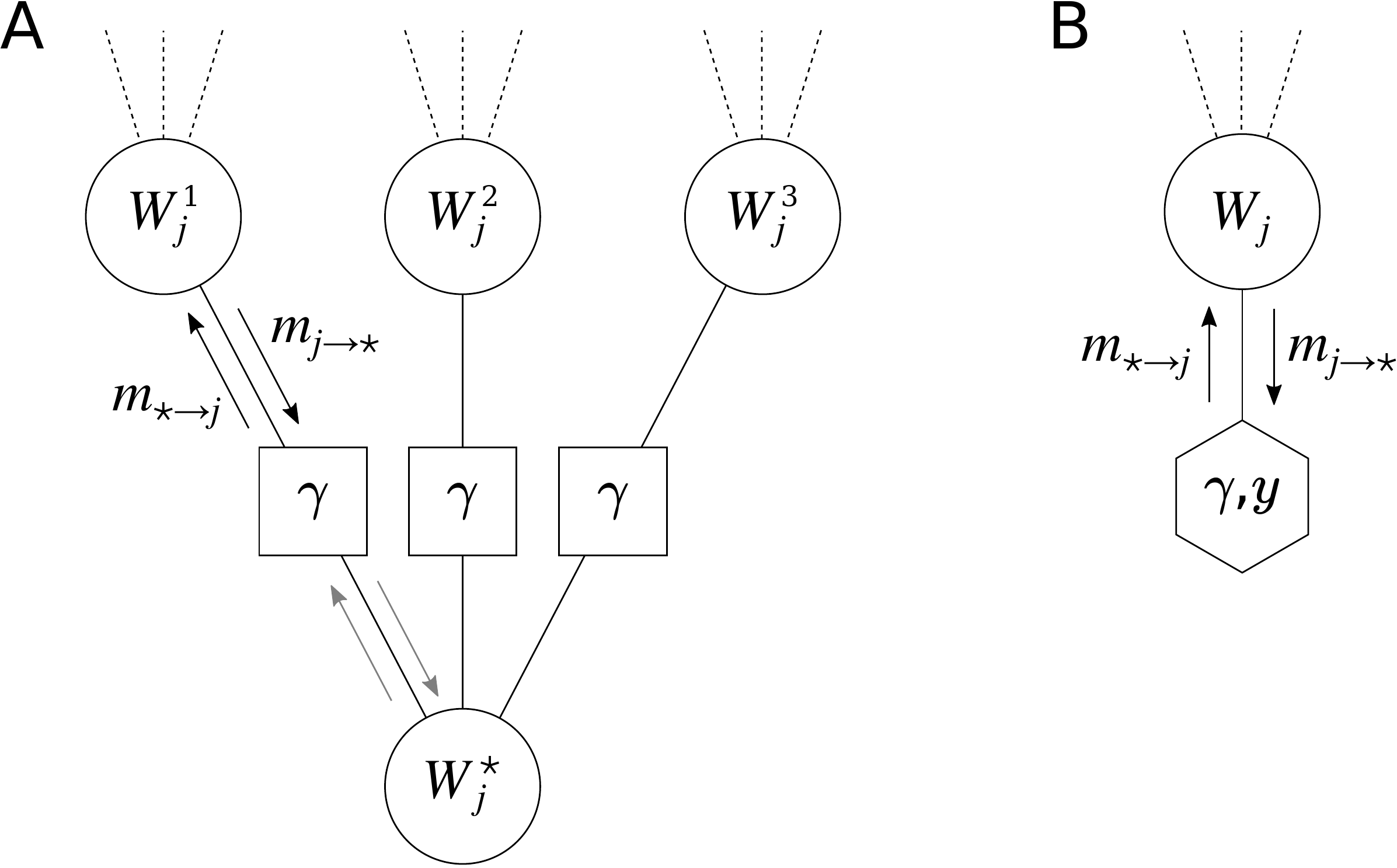}\caption{\textbf{\label{fig:BPR-scheme}A}. Portion of a BP factor graph for
a replicated variable $W_{j}$ with $y=3$ replicas and a reference
configuration $W_{j}^{\star}$. The dashed lines represent edges with
the rest of the factor graph. The squares represent the interactions
$\gamma W_{j}^{\star}W_{j}^{a}$. All BP messages (arrows) are assumed
to be the same in corresponding edges. \textbf{B}. Transformed graph
which represents the same graph as in A but exploits the symmetry
to reduce the number of nodes, keeping only one representative per
replica. The hexagon represents a pseudo-self-interaction, i.e.~it
expresses the fact that $m_{\star\to j}$ depends on $m_{j\to\star}$
and is parametrized by $\gamma$ and $y$.}
\end{figure}

Fortunately, it turns out that that there is a different way of applying
BP to the replicated system, leading to an efficient solver which
is both very similar to the reinforced BP algorithm and reasonably
well described by the theoretical results. Instead of considering
the joint distribution over all replicated variables at a certain
site $j$, we simply replicate the original factor graph $y$ times;
then, for each site $j$, we add an extra variable $W_{j}^{\star}$,
and $y$ interactions, between each variable $W_{j}^{a}$ and $W_{j}^{\star}$.
Finally, since the problem is symmetric, we assume that each replica
of the system behaves in exactly the same way, and therefore that
the same messages are exchanged along the edges of the graph regardless
of the replica index. This assumption neglects some of the correlations
between the replicated variables, but allows us to work only with
a single system, which is identical to the original one except that
each variable now also exchanges messages with $y-1$ identical copies
of itself through an auxiliary variable (which we can just trace away
at this point, as in eq.~(\ref{eq:part_func_traced})). The procedure
is shown graphically in Fig.~\ref{fig:BPR-scheme}. At each iteration
step $t$, each variable receives an extra message of the form: 
\begin{equation}
m_{\star\to j}^{t+1}=\tanh\left(\left(y-1\right)\tanh^{-1}\left(m_{j\to\star}^{t}\tanh\gamma\right)\right)\tanh\gamma\label{eq:pseudo-reinforcement}
\end{equation}
where $m_{j\to\star}^{t}$ is the cavity magnetization resulting from
the rest of the factor graph at time $t$, and $\gamma$ is the interaction
strength. This message simply enters as an additional term in the
sum in the right-hand side of eq.~(\ref{eq:BP}). Note that, even
though we started from a system of $y$ replicas, after the transformation
we are no longer constrained to keep $y$ in the integer domain. The
reinforced BP \cite{braunstein-zecchina}, in contrast, would have
a term of the form: 
\begin{equation}
m_{\star\to j}^{t+1}=\tanh\left(\rho\tanh^{-1}m_{j}^{t}\right)\label{eq:reinforcement}
\end{equation}
The latter equation uses a single parameter $0\le\rho\le1$ instead
of two, and is expressed in terms of the total magnetization $m_{j}^{t}$
instead of the cavity magnetization $m_{j\to\star}^{t}$. Despite
these differences, these two terms induce exactly the same BP fixed
points if we set $\gamma=\infty$ and $y=\left(1-\rho\right)^{-1}$
(see the Appendix section~\ref{app-sub:fBP-vs-rBP}). This result
is somewhat incidental, since in the context of our analysis the limit
$\gamma\to\infty$ should be reached gradually, otherwise the results
should become trivial (see figure~\ref{fig:entropic-landscape-sketch}),
and the reason why this does not happen when setting $\gamma=\infty$
in eq.~(\ref{eq:pseudo-reinforcement}) is only related to the approximations
introduced by the simplified scheme of figure~(\ref{fig:BPR-scheme}).
As it turns out, however, choosing slightly different mappings with
both $\gamma$ and $y$ diverging gradually (e.g.~$\gamma=\tanh^{-1}\sqrt{\rho}$
and $y=\frac{2-\rho}{1-\rho}$ with $\rho$ increasing from $0$ to
$1$) can lead to update rules with the same qualitative behavior
and very similar quantitative effects, such that the performances
of the resulting algorithms are hardly distinguishable, and such that
the aforementioned approximations do not thwart the consistency with
the theoretical analysis. This is shown and discussed Appendix~\ref{app-sub:fBP-vs-rBP}.
Using a protocol in which $\gamma$ is increased gradually, rather
than being set directly at $\infty$, also allows the algorithm to
reach a fixed point of the BP iterative scheme before proceeding to
the following step, which offers more control in terms of the comparison
with analytical results, as discussed in the next paragraph. In this
sense, we therefore have derived a qualitative explanation of the
effectiveness of reinforced BP within a more general scheme for the
search of accessible dense states. We call this algorithm Focusing
BP (fBP).\footnote{The code implementing fBP on Committee Machines with binary synaptic
weights is available at~\cite{codeFBP}.}

\begin{figure}
\includegraphics[width=0.9\textwidth]{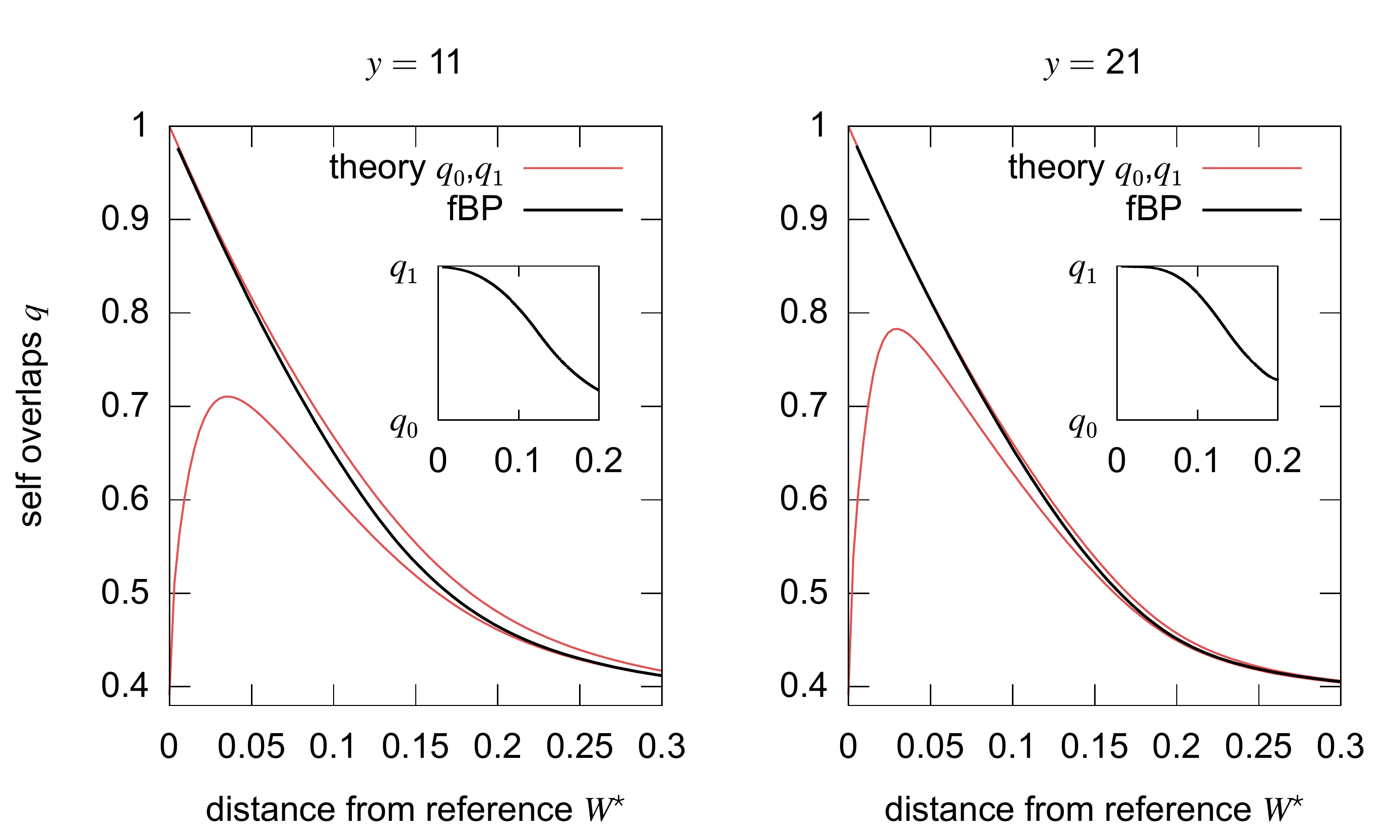}\caption{\label{fig:BP-symm-break}Focusing BP (fBP) spontaneously breaks replica
symmetry: the overlap order parameter $q$ (black thick curves) gradually
transitions from the inter-cluster overlap $q_{0}$ and the intra-cluster
overlap $q_{1}$ of the replica theory (red thin curves, $q_{0}<q_{1}$)
as the distance to the reference $W^{\star}$ goes to $0$ (i.e.~as
$\gamma\to\infty$). The insets provide an alternative visualization
of this phenomenon, plotting $\left(q-q_{0}\right)/\left(q_{1}-q_{0}\right)$
against the distance. These results were obtained on a perceptron
with $N=1001$ at $\alpha=0.6$, averaging over $50$ samples. The
two panels shows that the transition occurs at larger distances (i.e.~at
smaller $\gamma$) at larger $y$.}
\end{figure}

Apart from the possibility of using fBP as a solver, by gradually
increasing $\gamma$ and $y$ until a solution is found, it is also
interesting to compare its results at fixed values of $y$ and $\gamma$
with the analytical predictions for the perceptron case which were
derived in~\cite{baldassi_subdominant_2015,baldassi_local_2016},
since at the fixed point we can use the fBP messages to compute such
values as the local entropy, the distance from the reference configuration,
and the average overlap between replicas (defined as $q=\frac{1}{N}\sum_{j}\left\langle W_{j}^{a}\right\rangle \left\langle W_{j}^{b}\right\rangle $
for any $a$ and \textbf{$b$}, where $\left\langle \cdot\right\rangle $
denotes the thermal averaging). The expressions for these quantities
are provided in Appendix~\ref{app-sub:BP-notes}. The results indicate
that fBP is better than the alternatives described above in overcoming
the issues arising from RSB effects. The 1RSB scheme describes a non-ergodic
situation which arises from the break-up of the space of configurations
into a collection of several separate equivalent ergodic clusters,
each representing a possible state of the system. These states are
characterized by the typical overlap inside the same state (the intra-cluster
overlap $q_{1}$) and the typical overlap between configurations from
any two different states (the inter-cluster overlap $q_{0}$). Fig.~\textbf{\ref{fig:BP-symm-break}}
shows that the average overlap $q$ between replicas computed by the
fBP algorithm transitions from $q_{0}$ to $q_{1}$ when $\gamma$
is increased (and the distance with the reference is decreased as
a result). This suggests that the algorithm has spontaneously chosen
one of the possible states of high local entropy in the RE, achieving
an effect akin to the spontaneous symmetry breaking of the 1RSB description.
Within the state, replica symmetry holds, so that the algorithm is
able to eventually find a solution to the problem. Furthermore, the
resulting estimate of the local entropy is in very good agreement
with the 1RSB predictions up to at least $\alpha=0.6$ (see the Appendix,
figure~\ref{app-fig:BPpR-vs-replicas}).

Therefore, although this algorithm is not fully understood from the
theoretical point of view, it offers a valuable insight into the reason
for the effectiveness of adding a reinforcement term to the BP equations.
It is interesting in this context to observe that the existence of
a link between the reinforcement term and the robust ensemble seems
consistent with some recent results on random bicoloring constraint
satisfaction problems~\cite{braunstein_large_2016}, which showed
that reinforced BP finds solutions with shorter ``whitening times''
with respect to other solvers: this could be interpreted as meaning
they belong to a region of higher solution density, or are more central
in the cluster they belong to. Furthermore, our algorithm can be used
to estimate the point up to which accessible dense states exist, even
in cases, like multi-layer networks, where analytical calculations
are prohibitively complex.

\begin{figure}
\includegraphics[width=0.9\textwidth]{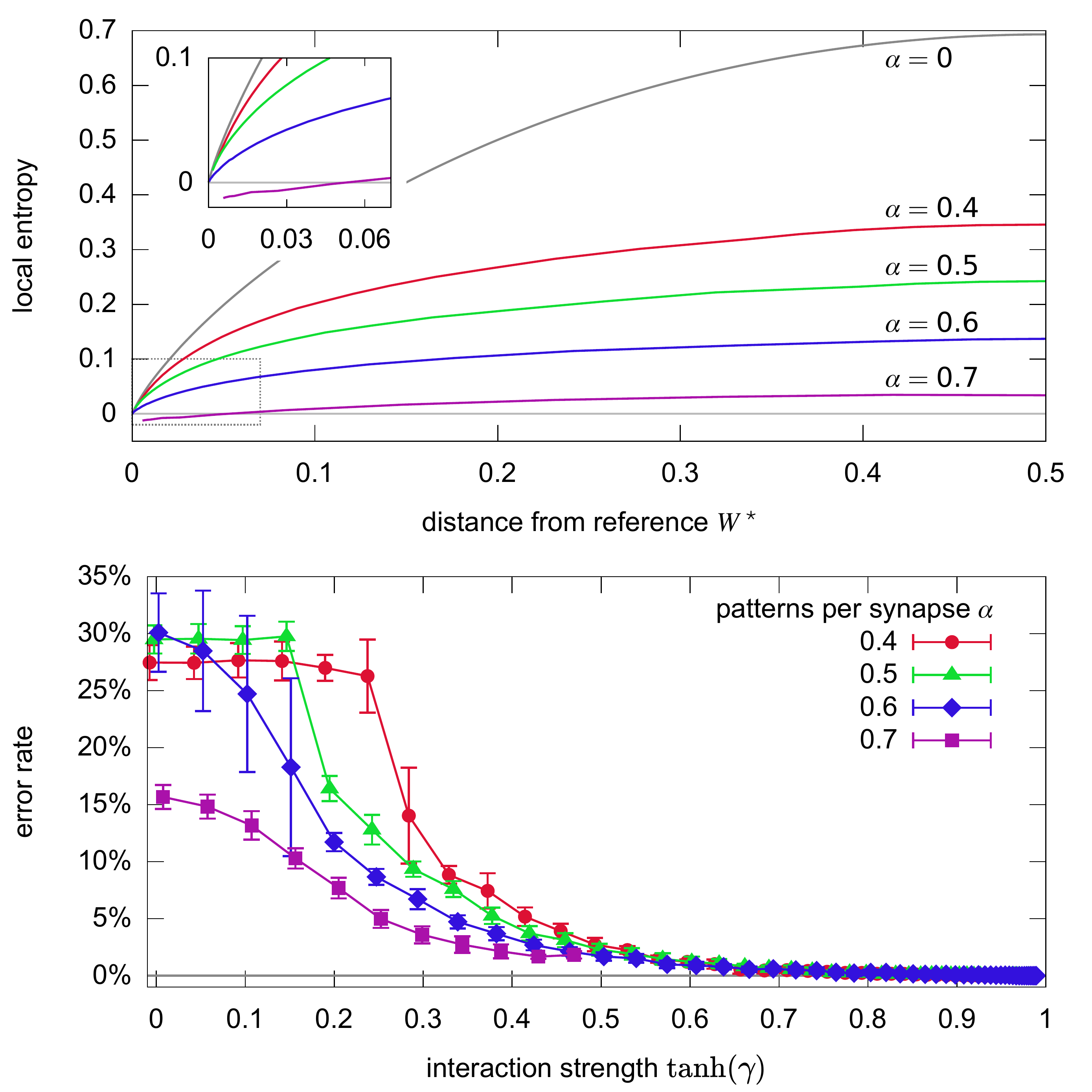}\caption{\label{fig:fBP-committee}Results of fBP on a committee machine with
$N=1605$, $K=5$, $y=7$, increasing $\gamma$ from $0$ to $2.5$,
averages on $10$ samples. Top: local entropy versus distance to the
reference $W^{\star}$ for various $\alpha$ (error bars not shown
for clarity). The topmost gray curve ($\alpha=0$) is an upper bound,
representing the case where all configurations within some distance
are solutions. Inset: enlargement of the region near the origin indicated
by the rectangle in the main plot. This shows that dense states exist
up to almost $\alpha=0.6$: at this value of $\alpha$, dense states
are only found for a subset of the samples (in which case a solution
is also found). Negative local entropies (curve at $\alpha=0.7$)
are unphysical, and fBP fails shortly after finding such values. Bottom:
error rates as a function of $\tanh\left(\gamma\right)$. For $\alpha\le0.6$,
all curves eventually get to $0$. However, only $7$ out of $10$
samples reached a sufficiently high $\gamma$ at $\alpha=0.6$, while
in $3$ cases the fBP equations failed. The curve for $\alpha=0.7$
is interrupted because fBP failed for all samples, in each case shortly
after reaching a negative local entropy. The plateaus at $\alpha=0.4$
and $\alpha=0.5$ are regions where the solution to the equations
are symmetric with respect to the permutation of the hidden units:
fBP spontaneously breaks that symmetry as well.}
\end{figure}

Fig.~\ref{fig:fBP-committee} shows the result of experiments performed
on a committee machine with the same architecture and same $y$ of
Fig.~\ref{fig:RSGD_R7_tau80}. The implementation closely follows~\cite{braunstein-zecchina}
with the addition of the self-interaction eq.~(\ref{eq:pseudo-reinforcement}),
except that great care is required to correctly estimate the local
entropy at large $\gamma$, due to numerical issues (see Appendix~\ref{app-sub:BP-notes}).
The figure shows that fBP finds that dense states (where the local
entropy curves approach the upper bound at small distances) exist
up to nearly $\alpha=0.6$ patterns per synapse, and that when it
finds those dense states it is correspondingly able to find a solution,
in perfect agreement with the results of the replicated gradient descent
algorithm.

Finally, we also performed some exploratory tests applying fBP on
the random $K$-satisfiability problem, and we found clear indications
that the performance of this algorithm is similar to that of Survey-Propagation-based
algorithms~\cite{mezard2002analytic,marino2015backtrackingSP}, although
a more detailed analysis is required to draw a more precise comparison..
The results are reported in Appendix~\ref{app-subsec:fBP-KSAT}.

\section{Discussion}

In this paper, we have presented a general scheme that can be used
to bias the search for low-energy configurations, enhancing the statistical
weight of large, accessible states. Although the underlying theoretical
description is based on a non-trivial large deviation measure, its
concrete implementation is very simple\textemdash replicate the system
and introduce an interaction between the replicas\textemdash and versatile,
in that it can be generally applied to a number of different optimization
algorithms or stochastic processes. We demonstrated this by applying
the method to Simulated Annealing, Gradient Descent and Belief Propagation,
but it is clear that the list of possible applications may be much
longer. The intuitive interpretation of the method is also quite straightforward:
a set of coupled systems is less likely to get trapped in narrow minima,
and will instead be attracted to wide regions of good (and mostly
equivalent) configurations, thus naturally implementing a kind of
robustness to details of the configurations.

The utility of this kind of search depends on the details of the problem
under study. Here we have mainly focused on the problem of training
neural networks, for a number of reasons. The first is that, at least
in the case of single-layer networks with discrete synapses, we had
analytical and numerical evidence that dense, accessible states exist
and are crucial for learning and improving the generalization performance,
and we could compare our findings with analytical results. The second
is that the general problem of training neural networks has been addressed
in recent years via a sort of collective search in the space of heuristics,
fueled by impressive results in practical applications and mainly
guided by intuition; heuristics are evaluated based on their effectiveness
in finding accessible states with good generalization properties.
It seems reasonable to describe these accessible states as regions
of high local entropy, i.e., wide, very robust energy minima: the
center of such a region can act as a Bayesian estimator for the whole
extensive neighborhood. Here we showed a simple way to exploit the
existence of such states efficiently, whatever the optimization algorithm
used. This not only sheds light on previously known algorithms, but
also suggests improvements or even entirely new algorithms. Further
work is required to determine whether the same type of phenomenon
that we observed here in simple models actually generalizes to the
deep and complex networks currently used in machine learning applications
(the performance boost obtained by the EASGD algorithm of~\cite{NIPS2015_5761}
being a first indication in this direction), and to investigate further
ways to improve the performance of learning algorithms, or to overcome
constraints (such as being limited to very low-precision computations).

It is also natural to consider other classes of problems in which
this analysis may be relevant. One application would be solving other
constraint satisfaction problems. For example, in~\cite{baldassi_local_2016}
we demonstrated that the EdMC algorithm can be successfully applied
to the random $K$-satisfiability problem, even though we had to resort
to a rough estimate of the local entropy due to replica symmetry breaking
effects. We have shown here that the fBP algorithm presented above
is also effective and efficient on the same problem. It is also interesting
to note here that large-deviation analyses\textemdash although different
from the one of the present paper\textemdash of a similar problem,
the random bicoloring constraint satisfaction problem,\cite{dall2008entropy,braunstein_large_2016}
have shown that atypical ``unfrozen'' solutions exist (and can be
found with the reinforced BP algorithm) well beyond the point in the
phase diagram where the overwhelming majority of solutions have become
``frozen''. Finally, an intriguing problem is the development of
a general scheme for a class of out-of-equilibrium processes attracted
to accessible states: even when describing a system which is unable
to reach equilibrium in the usual thermodynamic sense or is driven
by some stochastic perturbation, it is still likely that its stationary
state can be characterized by a large local entropy.
\begin{acknowledgments}
We wish to thank Y. LeCun and L. Bottou for encouragement and interesting
discussions about future directions for this work. CBa, CL and RZ
acknowledge the European Research Council for grant n\textdegree ~267915.
\end{acknowledgments}

\appendix

\section{Model and notation}

This Appendix text contains all the technical details of the algorithms
described in the main text, the techniques and the parameters we used
to obtain the results we reported. We also report some additional
results and report other minor technical considerations.

Preliminarily, we set a notation used throughout the rest of this
document which is slightly different from the one of the main text,
but more suitable for this technical description.

\subsection{The network model}

As described in the main text, we consider an ensemble of $y$ neural
networks with $K$ units and binary variables $W_{i}^{ka}\in\left\{ -1,1\right\} $
where $k\in\left\{ 1,\ldots,K\right\} $ is the unit index, $i\in\left\{ 1,\ldots,\nicefrac{N}{K}\right\} $
is the synaptic index and $a\in\left\{ 1,\ldots,y\right\} $ is the
replica index. Each network has thus $N$ synapses, where $N$ is
divisible by $K$. For simplicity, we assume both $K$ and $\nicefrac{N}{K}$
to be odd. The output of each unit is defined by a function $\tau\left(\xi;W\right)=\sign\left(\sum_{i=1}^{N/K}W_{i}\xi_{i}\right)$.
The output of the network is defined by a function $\zeta\left(\left\{ \xi^{k}\right\} _{k};\left\{ W^{k}\right\} _{k}\right)=\sign\left(\sum_{k=1}^{K}\tau\left(\xi^{k};W^{k}\right)\right)$
where $\xi^{k}$ represents the input to the $k$-th unit. In the
case $K=1$, this is equivalent to a single-layer network (also known
as perceptron). In the case where all $\xi^{k}$ are identical for
each $k$, this is equivalent to a fully-connected two-layer network
(also known as committee machine or consensus machine). If the $\xi^{k}$
are different for different values of $k$, this is a tree-like committee
machine. Note that, due to the binary constraint on the model, adding
weights to the second layer is redundant, since for all negative weights
in the second layer we could always flip both its weight and all the
weights of the unit connected to it. Therefore, without loss of generality,
we just set the weights of the second layer to $1$, resulting in
the above definition of the output function $\zeta$.

The scalar product between two replicas $a$ and $b$ is defined as
$W^{a}\cdot W^{b}=\sum_{k=1}^{K}\sum_{i=1}^{N/K}W_{i}^{ka}W_{i}^{kb}$.
For brevity of notation, in cases where the unit index does not play
a role, we will often just use a single index $j\in\left\{ 1,\ldots,N\right\} $,
e.g.~$W^{a}\cdot W^{b}=\sum_{j=1}^{N}W_{j}^{a}W_{j}^{b}$.

\subsection{Patterns}

The networks are trained on random input/output associations, i.e.~patterns,
$\left(\xi^{\mu},\sigma_{D}^{\mu}\right)$ where $\mu\in\left\{ 1,\ldots,\alpha N\right\} $
is the pattern index. The parameter $\alpha>0$ determines the load
of the network, so that the number of patterns is proportional to
the number of synapses. The inputs are binary vectors of $N$ elements
with entries $\xi_{i}^{k\mu}\in\left\{ -1,+1\right\} $, and the desired
outputs are also binary, $\sigma_{D}^{\mu}\in\left\{ -1,+1\right\} $.
Both the inputs and the outputs are extracted at random and are independent
and identically distributed (i.i.d.), except in the case of the fully-connected
committee machine where $\xi_{i}^{k\mu}=\xi_{i}^{k^{\prime}\mu}$
for all $k,k^{\prime}$ and therefore we only extract the values for
$k=1$.

We also actually exploit a symmetry in the problem and set all desired
outputs to $1$, since for each pattern its opposite must have an
opposite output, i.e.~we can always transform an input output pair
$\left(\xi^{\mu},\sigma_{D}^{\mu}\right)$ into $\left(\xi^{\mu\prime},1\right)$,
where the new pattern $\xi^{\mu\prime}=\sigma_{D}^{\mu}\xi^{\mu}$
has the same probability as $\xi^{\mu}$. 

\subsection{Energy definition\label{app-sub:Energy-definition}}

The energy, or cost, for each pattern is defined as the minimum number
of weights which need to be switched in order to correctly classify
the pattern, i.e.~in order to satisfy the relation $\zeta\left(\left\{ \xi^{k\mu}\right\} _{k},\left\{ W^{k}\right\} _{k}\right)=1$.
The total energy is the sum of the energies for all patterns, $E\left(W\right)=\sum_{\mu=1}^{\alpha N}E^{\mu}\left(W\right)$.

If the current configuration of the weights $W$ satisfies the pattern,
the corresponding energy is obviously $0$. Therefore, if the training
problem is satisfiable, the ground states with this energy definition
are the same as for the easier energy function given in terms of the
number of errors.

If the current configuration violates the pattern, the energy can
be computed as follows: we need to compute the minimum number $c^{\mu}$
of units of the first level which need to change their outputs, choose
the $c^{\mu}$ units which are easiest to fix, and for each of them
compute the minimum number of weights which need to be changed. In
formulas:

\begin{equation}
E^{\mu}\left(W\right)=\Theta\left(-\Delta_{\mathrm{out}}^{\mu}\right)\sum_{k=1}^{c^{\mu}}s_{k}^{\mu}\label{app-eq:energy}
\end{equation}
where:

\begin{eqnarray}
\Delta_{k}^{\mu} & = & \xi^{k\mu}\cdot W^{k}\label{app-eq:en-aux-1}\\
\Delta_{\mathrm{out}}^{\mu} & = & \sum_{k}\sign\left(\Delta_{k}^{\mu}\right)\\
s^{\mu} & = & \sort\left(\left\{ -\frac{1}{2}\left(\Delta_{k}^{\mu}-1\right),\;\forall k:\Delta_{k}^{\mu}<0\right\} \right)\\
c^{\mu} & = & \frac{1}{2}\left(-\Delta_{\mathrm{out}}^{\mu}+1\right)\label{app-eq:en-aux-4}
\end{eqnarray}
where the $\sort\left(\cdot\right)$ function returns its argument
sorted in ascending order. The above auxiliary quantities all depend
on $W$, but we omitted the dependency for clarity of notation.

In the single-layer case $K=1$ the expression simplifies considerably,
since $\Delta_{\mathrm{out}}^{\mu}=\xi^{\mu}\cdot W$ and reduces
to $E^{\mu}\left(W\right)=\Theta\left(-\Delta_{\mathrm{out}}^{\mu}\right)\frac{1}{2}\left(-\Delta_{\mathrm{out}}^{\mu}+1\right)$
.

\section{Replicated Simulated Annealing\label{app-sec:RSA}}

We run Simulated Annealing (plus ``scoping'') on a system of interacting
replicas. For simplicity, we trace away the reference configuration
which mediates the interaction. Thus, at any given step, we want to
sample from a probability distribution
\begin{eqnarray}
P\left(\left\{ W^{a}\right\} \right) & \propto & \sum_{W}\exp\left(-\beta\sum_{a=1}^{y}E\left(W^{a}\right)+\gamma\sum_{a=1}^{y}\sum_{j=1}^{N}W_{j}^{a}W_{j}\right)\nonumber \\
 & \propto & \exp\left(-\beta\sum_{a=1}^{y}E\left(W^{a}\right)+\sum_{j}\log\left(2\cosh\left(\gamma\sum_{a=1}^{y}W_{j}^{a}\right)\right)\right)\label{app-eq:prob_dist_traced}
\end{eqnarray}

The reference configuration is traced out in this representation,
but we can obtain its most probable value by just computing $\tilde{W}_{j}=\mathrm{sign}\sum_{a=1}^{y}W_{j}^{a}$.
It is often the case that, when the parameters are chosen appropriately,
$E\left(\tilde{W}\right)\le\left\langle E\left(W^{a}\right)\right\rangle _{a}$,
i.e.~that the energy of the center is lower than that of the group
of replicas. In fact, we found this to be a good rule-of-thumb criterion
to evaluate the choice of the parameters in the early stages of the
algorithmic process.

The most straightforward way to perform the sampling (at fixed $\beta$
and $\gamma$) is by using the Metropolis rule; the proposed move
is to flip a random synaptic weight from a random replica. Of course
the variation of the energy associated to the candidate move now includes
the interaction term, parametrized by $\gamma$, which introduces
a bias that favors movements in the direction of the center of mass
of the replicas.

We also developed an alternative rule for choosing the moves in a
biased way which implicitly accounts for the interaction term while
still obeying the detailed balance condition. This alternative rule
is generally valid in the presence of an external field and is detailed
at the end of this section. Its advantage consists in reducing the
rejection rate, but since the move proposal itself becomes more time
consuming it is best suited to systems in which computing the energy
cost of a move is expensive, so its usefulness depends on the details
of the model.

\subsection{Computing the energy shifts efficiently}

Here we show how to compute efficiently the quantity $E\left(W^{\prime}\right)-E\left(W\right)$
when $W^{\prime}$ and $W$ only differ in the value of one synaptic
weight $j$ and the energy is defined as in eq.~(\ref{app-eq:energy}).
To this end, we define some auxiliary quantities in addition to the
ones required for the energy computation, eqs.~(\ref{app-eq:en-aux-1})-(\ref{app-eq:en-aux-4})
(note that we omit the replica index $a$ here since this needs to
be done for each replica independently):
\begin{eqnarray}
P^{+} & = & \left\{ \mu:\Delta_{\mathrm{out}}^{\mu}=1\right\} \\
P^{-} & = & \left\{ \mu:\Delta_{\mathrm{out}}^{\mu}<0\right\} \\
\chi^{\mu} & = & \begin{cases}
1 & \textrm{if}\:s^{\mu}<0\wedge c^{\mu}<K\wedge s_{c^{\mu}}^{\mu}=s_{c^{\mu}+1}^{\mu}\\
0 & \mathrm{otherwise}
\end{cases}
\end{eqnarray}
 These quantities must be recomputed each time a move is accepted,
along with~(\ref{app-eq:en-aux-1})-(\ref{app-eq:en-aux-4}). Note
however that in later stages of the annealing process most moves are
rejected, and the energy shifts can be computed very efficiently as
we shall see below.

Preliminarily, we note that any single-flip move only affects the
energy contribution from patterns in $P^{+}\cup P^{-}$.

The contribution to the energy shift $\Delta E^{\mu}$ for a proposed
move $W_{i}^{k}\rightarrow-W_{i}^{k}$ is most easily written in pseudo-code:

\begin{algorithm}
\caption{Energy shift function $\Delta E^{\mu}\left(\mu,\,k,\,i,\,W_{i}^{k\mu}\right)$\label{app-alg:Energy-shift}}
  \uIf{$\mu \in P^+$}{
    \lIf{$\xi_i^\mu \ne W_i^{k\mu}$}{\Return{0}}\;
    \lIf{$\sign\left(\Delta_k^\mu\right) \ne 1$}{\Return{0}}\;
    \Return{1}\;
  }
  \uElseIf{$\mu \in P^-$}{
    \lIf{$\Delta_k^\mu > 1$}{\Return{0}}\;
    $d := -\xi_i^\mu W_i^{k\mu}$\;
    \lIf{$\Delta_k^\mu > 0 \wedge d = 1$}{\Return{0}}\;
    \lIf{$\Delta_k^\mu = 1$}{\Return{1}}\;
    $v := -(\Delta_k^\mu + 1)/2 + 1$\;
    \lIf{$v > s^\mu_{c^\mu}$}{\Return{0}}\;
    \lIf{$v < s^\mu_{c^\mu}$}{\Return{-d}}\;
    \lIf{$d = 1$}{\Return{-1}}\;
    \lIf{$\chi^\mu = 1$}{\Return{0}}\;
    \Return{1}\;
  }
  \Else{
    \Return{0}\;
  }
\end{algorithm}

Indeed, this function is greatly simplified in the single-layer case
$K=1$.

\subsection{Efficient Monte Carlo sampling}

Here we describe a Monte Carlo sampling method which is a modification
of the Metropolis rule when the system uses $N$ binary variables
$W_{j}$ and the Hamiltonian function can be written as:
\begin{equation}
H\left(W\right)=E\left(W\right)-\frac{1}{\beta}\sum_{j=1}^{N}k_{j}W_{j}
\end{equation}
where the external fields $k_{j}$ can only assume a finite (and much
smaller than $N$) set of values. The factor $\beta^{-1}$ is introduced
merely for convenience. Comparing this to eq.~(\ref{app-eq:prob_dist_traced}),
we see that, having chosen a replica index $a$ uniformly at random,
we can identify 
\begin{equation}
k_{j}=\frac{1}{2}\left(\log\left(\frac{\cosh\left(\gamma+\gamma\sum_{b\ne a}W_{j}^{b}\right)}{\cosh\left(-\gamma+\gamma\sum_{b\ne a}W_{j}^{b}\right)}\right)\right)\label{app-eq:k_j}
\end{equation}
.

Given a transition probability to go from state $W$ to state $W^{\prime}$,
$P\left(W\to W^{\prime}\right)$, the detailed balance equation reads:
\begin{equation}
P\left(W\right)P\left(W\to W^{\prime}\right)=P\left(W^{\prime}\right)P\left(W^{\prime}\to W\right)
\end{equation}

Let us split the transition explicitly in two steps: choosing the
index $j$ and accepting the move. The standard Metropolis rule is:
pick an index $j\in\left\{ 1,\dots,N\right\} $ uniformly at random,
propose the flip of $W_{j}$, accept it with probability $\min\left(1,e^{-\beta\Delta E_{W\to W^{\prime}}-2k_{j}W_{j}}\right)$,
where $\Delta E_{W\to W^{\prime}}=E\left(W^{\prime}\right)-E\left(W\right)$.
We want to reduce the rejection rate and incorporate the effect of
the field in the proposal instead. We write:
\begin{equation}
P\left(W\to W^{\prime}\right)=C\left(W\to W^{\prime}\right)A\left(W\to W^{\prime}\right)
\end{equation}
where $C$ is the choice of the index, and $A$ is the acceptance
of the move. Usually $C$ is uniform and we ignore it, but here instead
we try to use it to absorb the external field term in the probability
distribution. From detailed balance we have:
\begin{eqnarray}
\frac{A\left(W\to W^{\prime}\right)}{A\left(W^{\prime}\to W\right)} & = & \frac{p\left(W^{\prime}\right)}{p\left(W\right)}\frac{C\left(W^{\prime}\to W\right)}{C\left(W\to W^{\prime}\right)}\nonumber \\
 & = & e^{-\beta\Delta E_{W\to W^{\prime}}-2k_{j}W_{j}}\frac{C\left(W^{\prime}\to W\right)}{C\left(W\to W^{\prime}\right)}
\end{eqnarray}
so if we could satisfy:
\begin{equation}
e^{-2k_{j}W_{j}}\frac{C\left(W^{\prime}\to W\right)}{C\left(W\to W^{\prime}\right)}=1\label{app-eq:cond1}
\end{equation}
then the acceptance $A$ would simplify to the usual Metropolis rule,
involving only the energy shift $\Delta E$. This will turn out to
be impossible, yet easily fixable, so we still first derive the condition
implied by eq.~(\ref{app-eq:cond1}). The key observation is that
there is a finite number of classes of indices in $W$, based on the
limited number of values that $W_{j}k_{j}$ can take (in the case
of eq.~(\ref{app-eq:k_j}) there are $y$ possible values). Let us
call $K_{c}$ the possible classes, such that $W_{j}\in K_{c}\Leftrightarrow W_{j}k_{j}=c$,
and let us call $n_{c}=\left|K_{c}\right|$ their sizes, with the
normalization condition that $\sum_{c}n_{c}=N$. Within a class, we
must choose the move $j$ uniformly.

Then $C\left(W\to W^{\prime}\right)$ is determined by the probability
of picking a class, which in principle could be a function of all
the values of the $n_{c}$: $P_{c}\left(\left\{ n_{c^{\prime}}\right\} _{c^{\prime}}\right)$.
Suppose now that we have picked an index in a class $K_{c}$. The
transition to $W^{\prime}$ would bring it into class $K_{-c}$, and
the new class sizes would be
\[
n_{c^{\prime}}^{\prime}=\begin{cases}
n_{c^{\prime}}+1 & \mathrm{if}\:c^{\prime}=-c\\
n_{c^{\prime}}-1 & \mathrm{if}\:c^{\prime}=c\\
n_{c^{\prime}} & \mathrm{otherwise}
\end{cases}
\]
therefore: 
\begin{equation}
\frac{C\left(W^{\prime}\to W\right)}{C\left(W\to W^{\prime}\right)}=\frac{n_{c}}{P_{c}\left(\left\{ n_{c^{\prime}}\right\} _{c^{\prime}}\right)}\frac{P_{-c}\left(\left\{ n_{c^{\prime}}^{\prime}\right\} _{c^{\prime}}\right)}{n_{-c}+1}
\end{equation}

Since the only values of $n_{c^{\prime}}$ directly involved in this
expression are $n_{c}$ and $n_{-c}$, it seems reasonable to restrict
the dependence of $P_{c}$ and $P_{-c}$ only on those values. Let
us also call $q_{c}=n_{c}+n_{-c}$, which is unaffected by the transition.
Therefore we can just write:
\begin{equation}
\frac{C\left(W^{\prime}\to W\right)}{C\left(W\to W^{\prime}\right)}=\frac{n_{c}}{q_{c}-n_{c}+1}\frac{P_{-c}\left(q_{c}-n_{c}+1,q_{c}\right)}{P_{c}\left(n_{c},q_{c}\right)}
\end{equation}

Furthermore, we can assume \textendash{} purely for simplicity \textendash{}
that: 
\begin{equation}
P_{c}\left(n_{c},q_{c}\right)+P_{-c}\left(q_{c}-n_{c},q_{c}\right)=\frac{q_{c}}{N}\label{app-eq:assumption}
\end{equation}
which allows us to restrict ourselves in the following to the case
$c>0$, and which implies that the choice of the index will proceed
like this: we divide the indices in super-classes $D_{c}=K_{c}\cup K_{-c}$
of size $q_{c}$ and we choose one of those according to their size;
then we choose either the class $K_{c}$ or $K_{-c}$ according to
$P_{c}\left(n_{c},q_{c}\right)$; finally, we choose an index inside
the class uniformly at random. Considering this process, what we actually
need to determine is the conditional probability of choosing $K_{c}$
once we know we have chosen the super-class $D_{c}$:
\begin{equation}
\hat{P}_{c}\left(n_{c},q_{c}\right)=\frac{N}{q_{c}}P_{c}\left(n_{c},q_{c}\right)
\end{equation}

Looking at eq\@.~(\ref{app-eq:cond1}) we are thus left with the
condition:

\begin{equation}
\hat{P}_{c}\left(n_{c}+1,q_{c}\right)=e^{-2c}\frac{n_{c}+1}{q_{c}-n_{c}}\left(1-\hat{P}_{c}\left(n_{c},q_{c}\right)\right)\label{app-eq:recursive1}
\end{equation}

Considering that we must have $\hat{P}_{c}\left(0,q_{c}\right)=0$,
this expression allows us to compute recursively $\hat{P}_{c}\left(n_{c},q_{c}\right)$
for all values of $n_{c}$. The computation can be carried out analytically
and leads to $\hat{P}_{c}\left(n_{c},q_{c}\right)=\phi\left(n_{c},q_{c},e^{-2c}\right)$
where the function $\phi$ is defined as:
\begin{equation}
\phi\left(n,q,\lambda\right)=\lambda\frac{n}{q-n+1}\,{}_{2}F_{1}\left(1,1-n;q-n+2;\lambda\right)
\end{equation}
with $_{2}F_{1}$ the hypergeometric function. However, we should
also have $\hat{P}_{c}\left(q_{c},q_{c}\right)=1$, while $\phi\left(q,q,\lambda\right)=1-\left(1-\lambda\right)^{q}$
and therefore this condition is only satisfied for $c=0$ (in which
case we recover $\hat{P}_{c}\left(n_{c},q_{c}\right)=\frac{n_{c}}{q_{c}}$,
i.e.~the standard uniform distribution, as expected).

Therefore, as anticipated, eq.~(\ref{app-eq:cond1}) can not be satisfied\footnote{Strictly speaking we have not proven this, having made some assumptions
for simplicity. However it is easy to prove it in the special case
in which $k_{j}\in\left\{ -1,+1\right\} $, since then our assumptions
become necessary.}, and we are left with a residual rejection rate for the case $n_{c}=q_{c}$.
This is reasonable, since in the limit of very large $c$ (i.e.~very
large $\gamma$ in the case of eq.~(\ref{app-eq:k_j})) the probability
distribution of each spin must be extremely peaked on the state in
which all replicas are aligned, such that the combined probability
of all other states is lower than the probability of staying in the
same configuration. Therefore we have (still for $c>0$):
\begin{eqnarray}
\hat{P}_{c}\left(n_{c},q_{c}\right) & = & \phi\left(n_{c},q_{c},e^{-2c}\right)\left(1-\delta_{n_{c},q_{c}}\right)+\delta_{n_{c},q_{c}}\\
\frac{A\left(W\to W^{\prime}\right)}{A\left(W^{\prime}\to W\right)} & = & e^{-\beta\Delta E_{W\to W^{\prime}}}\left(1-\delta_{n_{c},q_{c}}\left(1-e^{-2c}\right)^{q_{c}}\right)
\end{eqnarray}
where $\delta_{n,q}$ is the Kronecker delta symbol. The last condition
can be satisfied by choosing a general acceptance rule of this form:
\begin{equation}
A\left(W\to W^{\prime}\right)=\min\left(1,e^{-\beta\Delta E_{W\to W^{\prime}}}\right)a_{c}\left(n_{c},q_{c}\right)
\end{equation}
where
\[
a_{c}\left(n_{c},q_{c}\right)=\begin{cases}
1-\delta_{n_{c},q_{c}}\left(1-e^{-2c}\right)^{q_{c}} & \mathrm{if}\:c>0\\
1 & \mathrm{if}\:c\le0
\end{cases}
\]

In practice, the effect of this correction is that the state where
all the variables in class $K_{c}$ are already aligned in their preferred
direction is a little ``clingier'' than the others, and introduces
an additional rejection rate $\left(1-e^{-2c}\right)^{q_{c}}$ (which
however is tiny when either $c$ is small or $q_{c}$ is large).

The final procedure is thus the following: we choose a super-class
$D_{c}$ at random with probability $\nicefrac{q_{c}}{N}$, then we
choose either $K_{c}$ or $K_{-c}$ according to $\hat{P}_{c}$ and
finally pick another index uniformly at random within the class.

This procedure is highly effective at reducing the rejection rate
induced by the external fields. As mentioned above, depending on the
problem, if the computation of the energy shifts is particularly fast,
it may still be convenient in terms of CPU time to produce values
uniformly and rejecting many of them, rather then go through a more
involved sampling procedure. Note however that the bookkeeping operations
required for keeping track of the classes compositions and their updates
can be performed efficiently, in $\mathcal{O}\left(1\right)$ time
with $\mathcal{O}\left(N\right)$ space, by using an unsorted partition
of the spin indices (which allows for efficient insertion/removal)
and an associated lookup table. Therefore, the additional cost of
this procedure is a constant factor at each iteration.

Also, the function $\phi\left(n,q,\lambda\right)$ involves the evaluation
of a hypergeometric function, which can be relatively costly; its
values however can be pre-computed and tabulated if the memory resources
allow it, since they are independent from the problem instance. For
large values of $q-n\left(1-\lambda\right)$, it can also be efficiently
approximated by a series expansion. It is convenient for that purpose
to change variables to
\begin{eqnarray*}
x & = & q-n\left(1-\lambda\right)\\
\rho & = & \frac{n\lambda}{x}
\end{eqnarray*}
(note that $\rho\in\left[0,1\right]$). We give here for reference
the expansion up to $x^{-2}$, which ensures a maximum error of $10^{-5}$
for $x\ge40$:
\begin{equation}
\phi\left(\frac{x\rho}{\lambda},x\left(1+\rho\frac{1-\lambda}{\lambda}\right),\lambda\right)=\rho\left(1-\frac{\left(1-\rho\right)\left(1-\lambda\right)}{x}\left(1+\frac{1-\left(2-3\rho\right)\left(1-\lambda\right)}{x}\left(1+\mathcal{O}\left(\frac{1}{x}\right)\right)\right)\right)
\end{equation}

Finally, note that the assumption of eq.~(\ref{app-eq:assumption})
is only justified by simplicity; it is likely that a different choice
could lead to a further improved dynamics.\footnote{We did in fact generalize and improve this scheme after the preparation
of this manuscript, see~\cite{baldassi2016rrrmc}.}

\subsection{Numerical simulations details\label{app-sub:RSA-simulations}}

Our Simulated Annealing procedure was performed as follows: we initialized
the replicated system in a random configuration, with all replicas
being initialized equally. The initial inverse temperature was set
to $\beta_{0}$, and the initial interaction strength to $\gamma_{0}$.
We then ran the Monte Carlo simulation, choosing a replica index at
random at each step and a synaptic index according to the modified
Metropolis rule described in the previous section, increasing both
$\beta$ and $\gamma$, by a factor $1+\beta_{f}$ and $1+\gamma_{f}$
respectively, for each $1000y$ accepted moves. The gradual increase
of $\beta$ is called `annealing' while the gradual increase of $\gamma$
is called `scoping'. Of course, since with our procedure the annealing/scoping
step is fixed, the quantities $\beta_{f}$ and $\gamma_{f}$ should
scale with $N$. The simulations are stopped as soon as any one of
the replicas reaches zero energy, or after $1000Ny$ consecutive non-improving
moves, where a move is classified as non-improving if it is rejected
by the Metropolis rule or it does not lower the energy (this definition
accounts for the situation where the system is trapped in a local
minimum with neighboring equivalent configurations at large $\beta$,
in which case the algorithm would keep accepting moves with $\Delta E=0$
without doing anything useful).

In order to compare our method with standard Simulated Annealing,
we just removed the interaction between replicas from the above described
case, i.e.~we set $\gamma_{0}=0$. This is therefore equivalent to
running $y$ independent (except for the starting configurations)
procedures in parallel, and stopping as soon as one of them reaches
a solution.

In order to determine the scaling of the solution time with $N$,
we followed the following procedure: for each sample (i.e.~patterns
assignment) we ran the algorithm with different parameters and recorded
the minimum number of iterations required to reach a solution. We
systematically explored these values of the parameters: $\beta_{0}\in\left\{ 0.1,0.5,1,2,3,\ldots,10\right\} $,
$\beta_{f}\in\left\{ 0.1,0.2,\ldots,4.9,5.0\right\} $, $\gamma_{0}\in\left\{ 0.1,0.5,1,1.5\right\} $,
$\gamma_{f}\in\left\{ 0,0.01,0.02,\ldots,0.4\right\} $ (the latter
two only in the interacting case, of course). This procedure gives
us an estimate for the minimum number of iterations required to solve
a typical problem at a given value of $N$, $K$ and $\alpha$. We
tested $10$ samples for each value of $\left(N,K,\alpha\right)$.
Since the interacting case has $2$ additional parameters, this implies
that there were more optimization opportunities, attributable to random
chance; this however is not remotely sufficient to explain the difference
in performance between the two cases: in fact, comparing instead for
the typical value of iterations required (i.e.~optimizing the average
iterations over $\left(\beta_{0},\beta_{f},\gamma_{0},\gamma_{f}\right)$)
gives qualitatively similar results, since once a range of good values
for the parameters is found the iterations required to reach a solution
are rather stable across samples.

\begin{figure}
\includegraphics[width=0.9\textwidth]{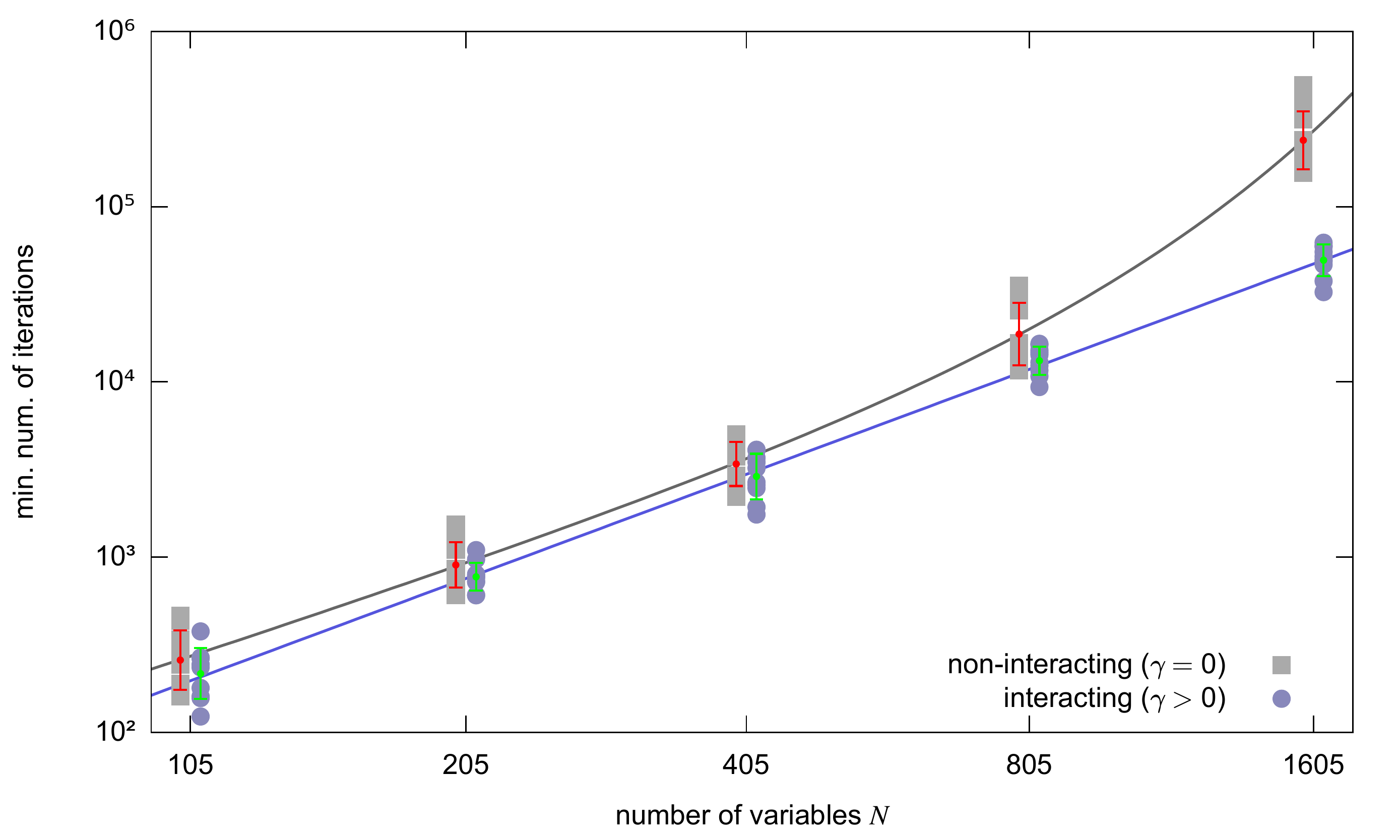}\caption{\label{app-fig:SimAnn}Replicated Simulated Annealing on the fully-connected
committee machine, with $K=5$ hidden units, comparison between the
interacting version (i.e.~which seeks regions of high solution density)
and the non-interacting version (i.e.~standard SA), at $\alpha=0.2$
using $y=3$ replicas. This is the analogous of figure~\ref{fig:SimAnnPerc}
of the main text for a committee machine, showing similar results.
$10$ samples were tested for each value of $N$ (the same samples
were used for the two curves). The bars represent averages and standard
deviations (taken in logarithmic scale) while the lines represent
fits. The interacting case was fitted by a function $aN^{b}$ with
$a\simeq0.02$, $b\simeq2.0$, while the non-interacting case was
fitted by a function $aN^{b}e^{cN^{d}}$ with $a\simeq0.08$, $b\simeq1.7$,
$c\simeq4.2\cdot10^{-5}$, $d\simeq1.5$. The two data sets are slightly
shifted relative to each other for presentation purposes.}

\end{figure}

The results are shown in figure~\ref{fig:SimAnnPerc} of the main
text for the single-layer case at $\alpha=0.3$ and figure~\ref{app-fig:SimAnn}
for the fully-connected two-layer case (committee machine) at $\alpha=0.2$
and $K=5$. In both cases we used $y=3$, which seems to provide good
results (we did not systematically explore different values of $y$).
The values of $\alpha$ were chosen so that the standard SA procedure
would be able to solve some instances at low $N$ in reasonable times
(since the difference in performance between the interacting and non-interacting
cases widens greatly with increasing $\alpha$). The results show
a different qualitative behavior in both cases, polynomial for the
interacting case and exponential for the non-interacting cases. All
fits were performed directly in logarithmic scale. A similar behavior
is observed for the tree-like committee machine (not shown).

\section{Replicated Gradient Descent\label{app-sec:RGD}}

\subsection{Gradient computation}

As mentioned in the main text, we perform a stochastic gradient descent
on binary networks using the energy function of eq.~(\ref{app-eq:energy})
by using two sets of variables: a set of continuous variables $\mathcal{W}_{i}^{k}$
and the corresponding binarized variables $W_{i}^{k}$, related by
$W_{i}^{k}=\sign\left(\mathcal{W}_{i}^{k}\right)$. We use the binarized
variables to compute the energy and the gradient, and apply the gradient
to the continuous variables. In formulas, the quantities at time $t+1$
are related to those at time $t$ by:
\begin{eqnarray}
\left(\mathcal{W}_{i}^{k}\right)^{t+1} & = & \left(\mathcal{W}_{i}^{k}\right)^{t}-\eta\frac{1}{\left|m\left(t\right)\right|}\sum_{\mu\in m\left(t\right)}\frac{\partial}{\partial W_{i}^{k}}E^{\mu}\left(W^{t}\right)\label{app-eq:gdesc_noninteract}\\
\left(W_{i}^{k}\right)^{t+1} & = & \sign\left(\left(\mathcal{W}_{i}^{k}\right)^{t+1}\right)\label{app-eq:discr}
\end{eqnarray}
where $\eta$ is a learning rate and $m\left(t\right)$ is a set of
pattern indices (a so-called minibatch). A particularly simple scenario
can be obtained by considering a single layer network without replication
($K=1$, $y=1$) and a fixed learning rate, and by computing the gradient
one pattern at a time ($\left|m\left(t\right)\right|=1$). In that
case, $E^{\mu}\left(W\right)=R\left(-\sum_{i}W_{i}\xi_{i}^{\mu}\right)$
where $R\left(x\right)=\frac{1}{2}\left(x+1\right)\Theta\left(x\right)$
and the gradient is $\partial_{W_{i}}E^{\mu}\left(W\right)=-\frac{1}{2}\xi_{i}^{\mu}\Theta\left(-\sum_{i}W_{i}\xi_{i}^{\mu}\right)$.
Since the relation~(\ref{app-eq:discr}) is scale-invariant, we can
just set $\eta=4$ and obtain
\begin{eqnarray}
\mathcal{W}_{i}^{t+1} & = & \mathcal{W}_{i}^{t}-2\xi_{i}^{\mu}\Theta\left(-\sum_{i}W_{i}^{t}\xi_{i}^{\mu}\right)\label{app-eq:CP}
\end{eqnarray}
where now the auxiliary quantities $\mathcal{W}$ are discretized:
if they are initialized as odd integers, they remain odd integers
throughout the learning process. This is the so-called ``Clipped
Perceptron'' (CP) rule, which is the same as the Perceptron rule
(``in case of error, update the weights in the direction of the pattern,
otherwise do nothing'') except that the weights are clipped upon
usage to make them binary. Notably, the CP rule by itself does not
scale well with $N$; it is however possible to make it efficient
(see~\cite{baldassi-et-all-pnas,baldassi-2009}).

In the two-layer case ($K>1$) the computation of the gradient is
more complicated; it is however simpler than the computation of the
energy shift which was necessary for Simulated Annealing (Algorithm~\ref{app-alg:Energy-shift}),
since we only consider infinitesimal variations when computing the
gradient. The resulting expression is:
\begin{equation}
\partial_{W_{i}^{k}}E^{\mu}\left(W\right)=\begin{cases}
-\frac{1}{2}\xi_{i}^{k\mu} & \mathrm{if\:}\left(\Delta_{\mathrm{out}}^{\mu}<0\right)\wedge\left(1+2s_{c^{\mu}}^{\mu}\le\Delta_{k}^{\mu}<0\right)\\
0 & \mathrm{otherwise}
\end{cases}
\end{equation}
i.e.~the gradient is non-zero only in case of error, and only for
those units $k$ which contribute to the energy computation (which
turn up in the first $c^{\mu}$ terms of the sorted vector $s^{\mu}$,
see eqs.~(\ref{app-eq:en-aux-1})-(\ref{app-eq:en-aux-4})). Again,
since this gradient can take only $3$ possible values, we could set
$\eta=4$ and use discretized odd variables for the $\mathcal{W}$.

It is interesting to point out that a slight variation of this update
rule in which only the first, least-wrong unit is affected, i.e.~in
which the condition $\left(1+2s_{c^{\mu}}^{\mu}\le\Delta_{k}^{\mu}\right)$
is changed to $\left(1+2s_{1}^{\mu}\le\Delta_{k}^{\mu}\right)$, was
used in~\cite{baldassi_subdominant_2015}, giving good results on
a real-world learning task when a slight modification analogous to
the one of~\cite{baldassi-2009} was added. Note that, in the later
stages of learning, when the overall energy is low, it is very likely
that $c^{\mu}\le1$, implying that the simplification used in~\cite{baldassi_subdominant_2015}
likely has a negligible effect. The simplified version, when used
in the continuous case, also goes under the name of ``least action''
algorithm \cite{mitchison1989bounds}.

Having computed the gradient of $E\left(W\right)$ for each system,
the extension to the replicated system is rather straightforward,
since the energy (with the traced-out center) becomes (cf.~eqs.~(\ref{eq:part_func_traced})
and~(\ref{eq:interaction_traced}) in the main text):
\begin{equation}
H\left(\left\{ W^{a}\right\} \right)=\sum_{a=1}^{y}E\left(W^{a}\right)-\frac{1}{\beta}\sum_{j=1}^{N}\log\left(e^{-\frac{\gamma}{2}\sum_{a=1}^{y}\left(W_{j}^{a}-1\right)^{2}}+e^{-\frac{\gamma}{2}\sum_{a=1}^{y}\left(W_{j}^{a}+1\right)^{2}}\right)\label{app-eq:traced_hamiltonian}
\end{equation}
and therefore the gradient just has an additional term: 
\begin{equation}
\frac{\partial H}{\partial W_{i}^{a}}\left(\left\{ W^{b}\right\} \right)=\left.\frac{\partial E}{\partial W_{i}}\left(W\right)\right|_{W=W^{a}}-\frac{\gamma}{\beta}\left(\tanh\left(\gamma\sum_{b=1}^{y}W_{i}^{b}\right)-W_{i}^{a}\right)
\end{equation}

Note that the trace operation brings the parameter $\beta$ into account.
Using $\eta^{\prime}=\frac{\gamma}{\beta\eta}$ as control parameter,
the update equation~(\ref{app-eq:gdesc_noninteract}) for a replica
$a$ becomes (we omit the unit index $k$ for simplicity):
\begin{eqnarray}
\left(\mathcal{W}_{i}^{a}\right)^{t+1} & = & \left(\mathcal{W}_{i}^{a}\right)^{t}-\eta\frac{1}{\left|m\left(t\right)\right|}\sum_{\mu\in m\left(t\right)}\left.\frac{\partial E^{\mu}}{\partial W_{i}}\left(W\right)\right|_{W=\left(W^{a}\right)^{t}}+\eta^{\prime}\left(\tanh\left(\gamma\sum_{b=1}^{y}\left(W_{i}^{b}\right)^{t}\right)-\left(W_{i}^{a}\right)^{t}\right)\label{app-eq:SGDR}
\end{eqnarray}

In the limit $\beta,\gamma\to\infty$, $\eta^{\prime}$ stays finite,
while the $\tanh$ reduces to a $\sign$.

The expression of eq.~(\ref{app-eq:SGDR}) is derived straightforwardly,
gives good results and is the one that we have used in the tests shown
in the main text and below. It could be noted, however, that the two-level
precision of the variables used in the algorithm introduces some artifacts.
As a clear example, in the case of a single replica ($y=1$) or, more
in general, when the replica indices $W_{i}^{a}$ are all aligned,
we would expect the interaction term to vanish, while this is not
the case except at $\gamma=\infty$.

One possible way to fix this issue is the following: we can introduce
a factor in the logarithm in expression~(\ref{app-eq:traced_hamiltonian}):
\begin{equation}
\log\left(\frac{e^{-\frac{\gamma}{2}\sum_{a=1}^{y}\left(W_{j}^{a}-1\right)^{2}}+e^{-\frac{\gamma}{2}\sum_{a=1}^{y}\left(W_{j}^{a}+1\right)^{2}}}{f\left(W_{j}^{1},\dots,W_{j}^{y}\right)}\right)
\end{equation}
such that $f\left(W_{j}^{1},\dots,W_{j}^{y}\right)=1$ whenever its
arguments lie on the vertices of the hypercube, $W_{j}^{a}\in\left\{ -1,1\right\} $.
This does not change the Hamiltonian for the configurations we're
interested in, but it can change its gradient. We can thus impose
the additional constraint that the derivative of the above term vanishes
whenever the $W_{j}^{a}$ are all equal. There are several ways to
achieve this; however, if we assume that the function $f$ has the
general structure
\begin{equation}
f\left(W_{j}^{1},\dots,W_{j}^{y}\right)=a\left(g\left(W_{j}^{1}\right),\dots,g\left(W_{j}^{y}\right)\right)
\end{equation}
 with $g\left(1\right)=g\left(-1\right)$ and with $a\left(\dots\right)$
being a totally symmetric function of its arguments\footnote{One possibility is using $g\left(w\right)=\frac{\cosh\left(\gamma yw\right)}{\cosh\left(\gamma y\right)}\exp\left(\frac{\gamma y}{2}\left(1-w^{2}\right)\right)$
and $a\left(\dots\right)$ equal to the average of its arguments.}, then it can be easily shown that necessarily
\begin{equation}
\frac{\partial}{\partial W_{i}^{a}}\log\left(\frac{e^{-\frac{\gamma}{2}\sum_{b=1}^{y}\left(W_{j}^{b}-1\right)^{2}}+e^{-\frac{\gamma}{2}\sum_{b=1}^{y}\left(W_{j}^{b}+1\right)^{2}}}{a\left(g\left(W_{j}^{1}\right),\dots,g\left(W_{j}^{y}\right)\right)}\right)=\gamma\left(\tanh\left(\gamma\sum_{b=1}^{y}W_{i}^{b}\right)-\tanh\left(\gamma y\right)W_{i}^{a}\right)
\end{equation}

This expression has now two zeros corresponding to the fully aligned
configurations of the weights at $+1$ and $-1$, as desired, and
is a very minor correction of the original one used in eq.~(\ref{app-eq:SGDR})
(the expressions become identical at large values of $\gamma y$).
In fact, we found that the numerical results are basically the same
(the optimal values of the parameters may change, but the performances
for optimal parameters are very similar for the two cases), such that
this correction is not needed in practice.

An alternative, more straightforward way to fix the issue of the non-vanishing
gradient with aligned variables is to perform the trace over the reference
configurations in the continuous case (i.e.~replacing the sum over
the binary hypercube with an integral). This leads to the an expression
for the interaction contribution to the gradient of this form: $\gamma\left(\frac{1}{y}\sum_{b=1}^{y}W_{i}^{b}-W_{i}^{a}\right)$.
This, however, does seem to have a very slightly but measurably worse
overall performance with respect to the previous ones (while still
dramatically outperforming the non-interacting version).

In general, the tests with alternative interaction terms show that,
despite the fact that the two-level gradient procedure is purely heuristic
and inherently problematic, the fine details of the implementation
may not be exceedingly relevant for most practical purposes.

The code for our implementation is available at~\cite{codeRSGD}.

\subsection{Numerical simulations details}

Our implementation of the formula in eq.~(\ref{app-eq:SGDR}) follows
this scheme: at each time step, we have the values $T_{i}=\sum_{b=1}^{y}W_{i}^{b}$,
we pick a random replica index $a$, compute the gradient with respect
to some $m\left(t\right)$ patterns, update the values $\mathcal{W}^{a}$
and $W^{a}$, compute the gradient with respect to the interaction
term using $T$ and $W^{a}$, and update the values of $T$ and \textendash{}
again \textendash{} of $\mathcal{W}^{a}$ and $W^{a}$. This scheme
is thus easy to parallelize, since it alternates the standard learning
periods in which each replica acts independently with brief interaction
periods in which the sum $T$ is updated, similarly to what was done
in~\cite{NIPS2015_5761}.

An epoch consists of a presentation of all patterns to all replicas.
The minibatches $m\left(t\right)$ are randomized at the beginning
of each epoch, independently for each replica. The replicas were initialized
equally for simplicity.

In our tests, we kept fixed the learning rates $\eta$ and $\eta^{\prime}$
during the training process, since preliminary tests did not show
a benefit in adapting them dynamically in our setting. We did, however,
find beneficial in most cases to vary $\gamma,$ starting at some
value $\gamma_{0}$ and increasing it progressively by adding a fixed
quantity $d\gamma$ after each epoch, i.e.~implementing a ``scoping''
mechanism as in the Simulated Annealing case (although even just using
$\gamma=\infty$ from the start already gives large improvements against
the non-interacting version).

All tests were capped at a maximum of $10^{4}$ epochs, and the minimum
value of the error across all replicas was kept for producing the
graphs.

In the interacting case, we systematically tested various values of
$\eta^{\prime}$, $\gamma_{0}$ and $d\gamma$, and, for each $\alpha$,
we kept the ones which produced optimal results (i.e.~lowest error
rate, or shorter solution times if the error rates were equal) on
average across the samples. Because of the overall scale invariance
of the problem, we did not change $\eta$.

\begin{figure}
\includegraphics[width=0.95\columnwidth]{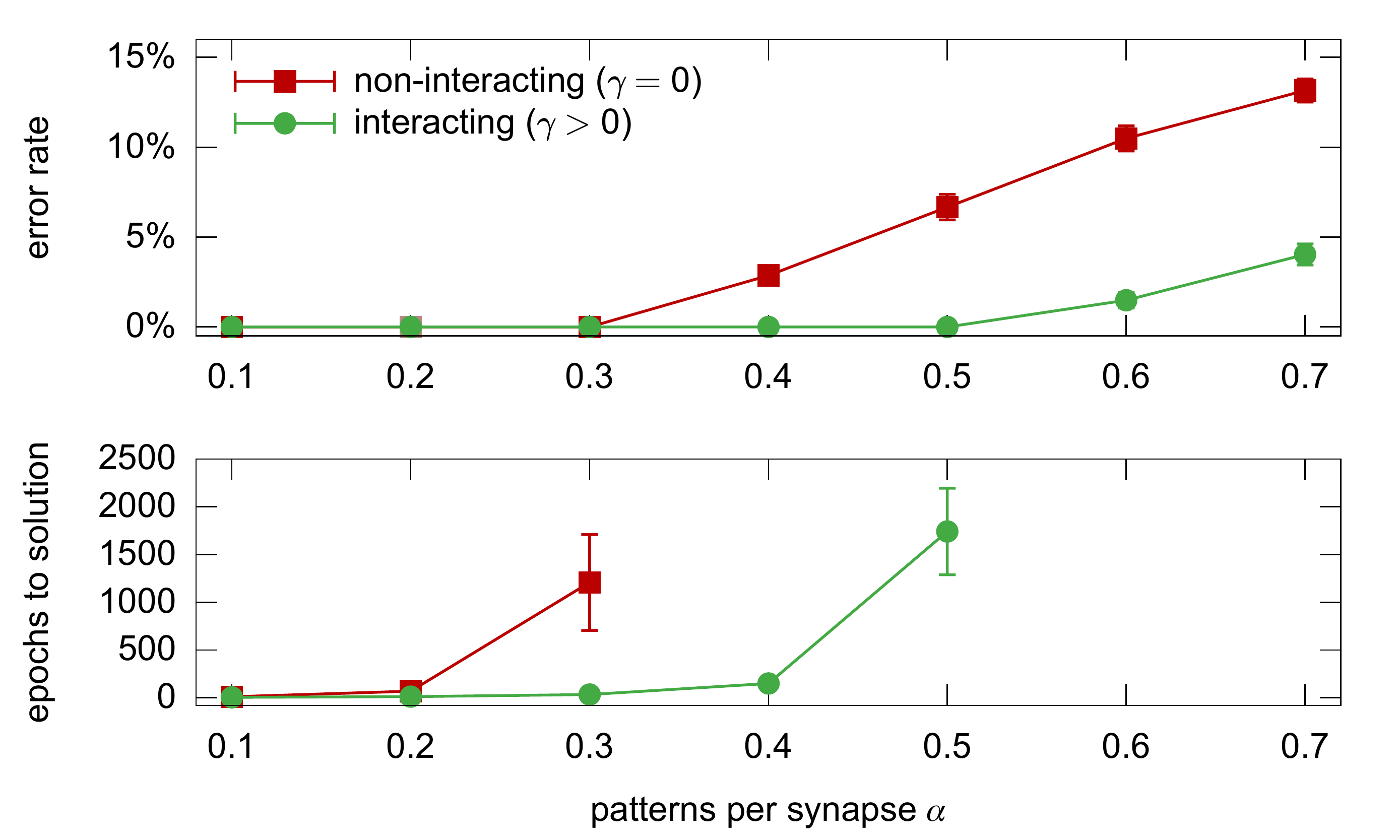}\caption{\label{app-fig:RSGD_R3_tau10}Replicated Stochastic Gradient descent
on a fully-connected committee machine with $N=1605$ synapses and
$K=5$ units in the second layer, comparison between the non-interacting
(i.e.~standard SGD) and interacting versions, using $y=3$ replicas
and a minibatch size of $10$ patterns. Each point shows averages
and standard deviations on $10$ samples with optimal choice of the
parameters, as a function of the training set size. Top: minimum training
error rate achieved after $10^{4}$ epochs. Bottom: number of epochs
required to find a solution. Only the cases with $100\%$ success
rate are shown.}

\end{figure}

Figure~\ref{app-fig:RSGD_R3_tau10} shows the results of the same
tests as shown in figure~\ref{fig:RSGD_R7_tau80} of the main text
for different values of the number of replicas and the minibatch size.
The results for the interacting case are slightly worse, but still
much better than for the non-interacting case.

For the perceptron case $K=1$ we did less extensive tests; we could
solve $10$ samples out of $10$ with $N=1601$ synapses at $\alpha=0.7$
in an average of $4371\pm661$ epochs with $y=7$ replicas and a minibatch
size of $80$ patterns.

\section{Replicated Belief Propagation}

\subsection{Belief Propagation implementation notes\label{app-sub:BP-notes}}

Belief Propagation (BP) is an iterative message passing algorithm
that can be used to derive marginal probabilities on a system within
the Bethe-Peierls approximation~\cite{mackay2003information,yedidia2005constructing,mezard_information_2009}.
The messages $P_{j\to\mu}\left(\sigma_{j}\right)$ (from variable
node $j$ to factor node $\mu$) and $P_{\mu\to j}\left(\sigma_{j}\right)$
(from factor node $a$ to variable node $j$) represent cavity probability
distributions (called messages) over a single variable $\sigma_{j}$.
In the case of Ising systems of binary $\pm1$ variables like the
ones we are using in the network models considered in this work, the
messages can be represented as a single number, usually a magnetization
$m_{i\to\mu}=P_{i\to\mu}\left(+1\right)-P_{\mu\to i}\left(-1\right)$
(and analogous for the other case).

\begin{figure}

\subfloat[\label{app-fig:BP-factor-graph}BP factor graph scheme. This scheme
exemplifies a factor graph for a committee machine with $N=15$ variables,
$K=3$ units in the second layer, trained on $2$ patterns. The two
patterns are distinguished by different colors. The graph can represent
a fully-connected committee machine if the patterns are the same for
all first-layer units, or a tree-like one if they are different. The
variable nodes are represented as circles, the interaction by other
geometrical figures. The hexagons at the bottom represent pseudo-self-interaction
nodes (see main text, figure~\ref{fig:BPR-scheme}), the large squares
with rounded corners represent perceptron-like nodes, the small squares
at the top represent external fields enforcing the desired output
of the machine. The synaptic variables $W_{j}^{k}$ are at the bottom
(black circles), while the rest of the variables are auxiliary and
represent the output of each unit for a given pattern.]{\includegraphics[width=0.48\columnwidth]{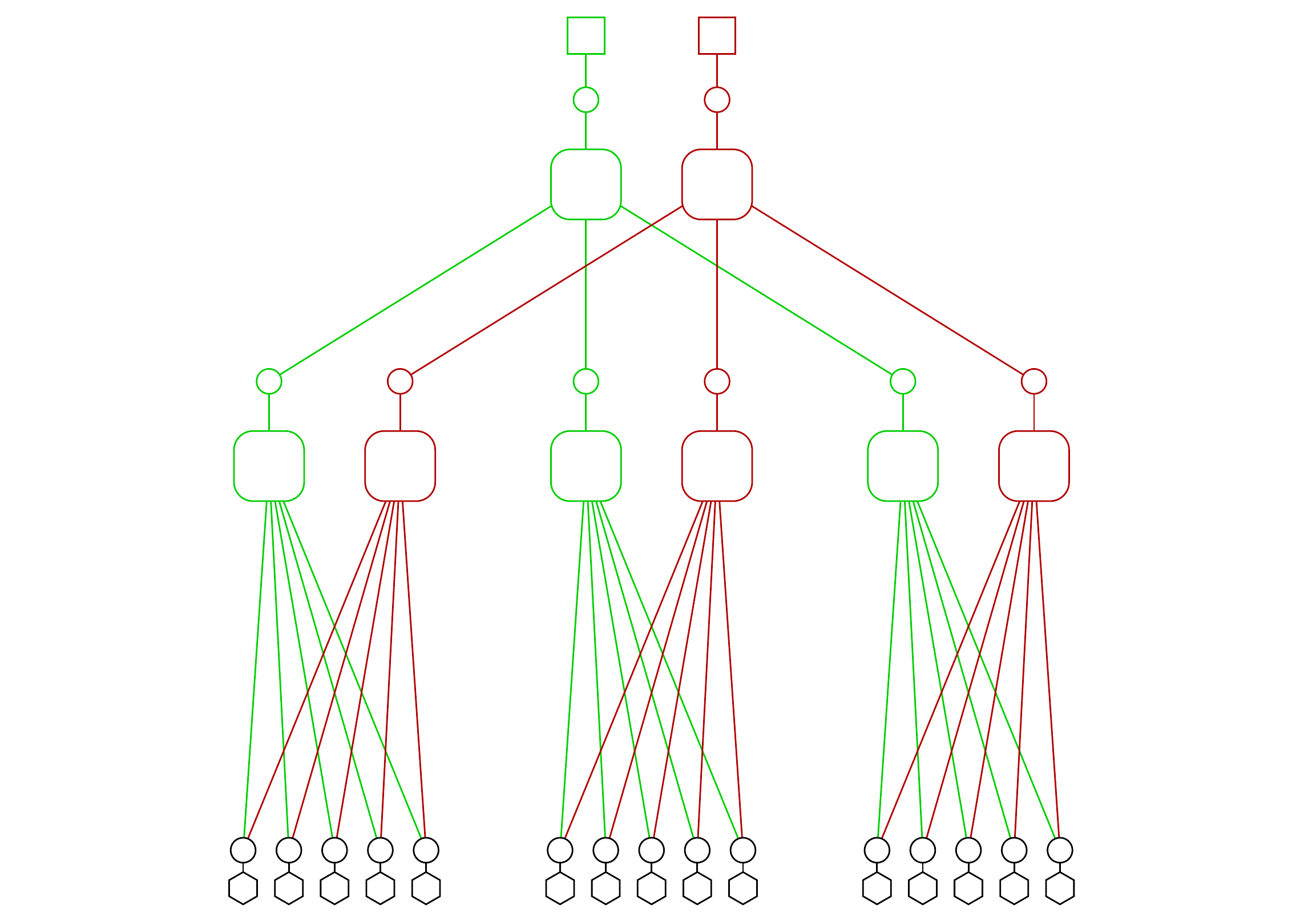}

}\hspace*{\fill}\subfloat[\label{app-fig:BP-perc-node}BP messages naming scheme used in section~\ref{app-sub:BP-notes}
for a perceptron-like factor node. The node $\mu$ is represented
by the central square. Input variables, denoted by $\sigma_{j}$,
are at the bottom. The output variable is called $\tau$. The couplings
$\xi_{j}^{\mu}$ parametrize the factor node (one parameter per input
edge) and can either represent an input pattern (for the first layer
of the network) or be $1$ (for the second layer of the network).]{\includegraphics[width=0.48\columnwidth]{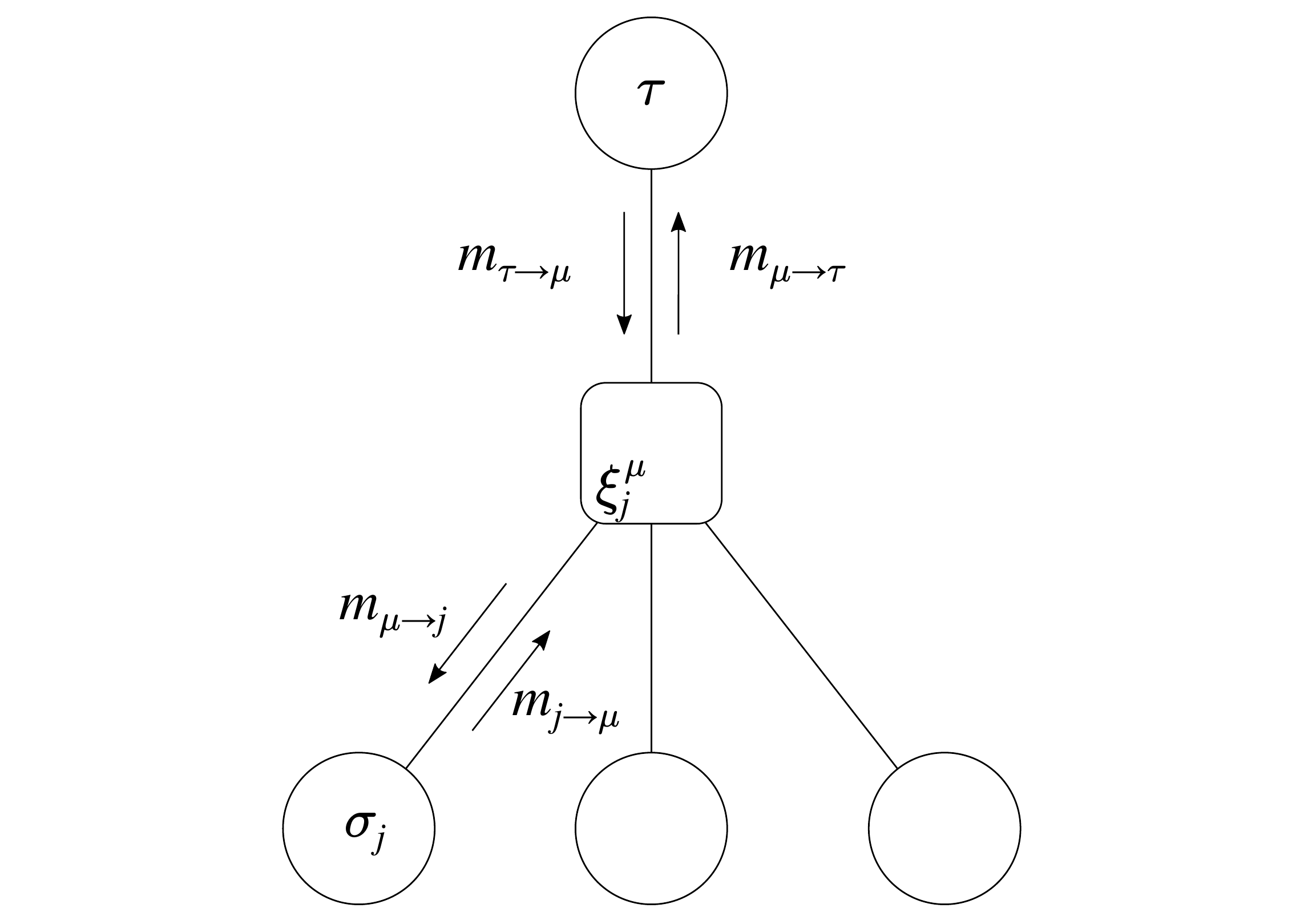}

}\caption{}
\end{figure}

Our implementation of BP on binary networks follows very closely that
of~\cite{braunstein-zecchina}, since we only consider the zero temperature
case and we are interested in the ``satisfiable'' phase, thus considering
only configurations of zero energy. However, in order to avoid some
numerical precision issues that affected the computations at high
values of $\alpha$, $y$ and $\gamma$, we lifted some of the approximations
used in that paper. Here therefore we recapitulate the BP equations
used and highlight the differences with the previous work. The factor
graph scheme for a committee machine is shown for reference in figure~\ref{app-fig:BP-factor-graph}.
The BP equations for the messages from a variable node $j$ to a factor
node $\mu$ can be written in general as:
\begin{eqnarray}
m_{j\to\mu}^{t} & = & \tanh\left(\sum_{\nu\in\partial j\setminus\mu}\tanh^{-1}\left(m_{\nu\to j}^{t}\right)+\tanh^{-1}\left(m_{\star\to j}^{t}\right)\right)
\end{eqnarray}
where $\partial j$ represent the set of all factor nodes in which
variable $j$ is involved. This expression includes the Focusing BP
(fBP) extra message $m_{\star\to j}$ described in the main text.
The general expression for perceptron-like factor nodes is considerably
more complicated. For the sake of generality, here we will use the
symbol $\sigma$ to denote input variables of the node (with subscript
$j$ indicating the variable), and $\tau$ for the output variable.
To each perceptron-like factor $\mu$ is associated a vector of couplings
$\xi^{\mu}$: in a committee-machine, these represent the patterns
for the first layer nodes, and are simply vectors of ones in the second
layer. See figure~\ref{app-fig:BP-perc-node}.

Let us define the auxiliary functions:
\begin{eqnarray}
\mathrm{f}_{j}^{\mu}\left(\left\{ m_{i\to\mu}\right\} _{i\in\partial\mu\setminus j},m_{\tau\to\mu},\sigma_{j}\right) & = & \sum_{\tau,\sigma_{\partial\mu\setminus j}}\!\left(\frac{1+\tau\,m_{\tau\to\mu}}{2}\right)\Theta\negthinspace\left(\tau\left(\sum_{i\in\partial\mu\setminus j}\xi_{i}^{\mu}\sigma_{i}+\xi_{j}^{\mu}\sigma_{j}\right)\!\!\right)\!\prod_{i\in\partial\mu\setminus j}\left(\frac{1+\sigma_{i}m_{i\to\mu}}{2}\right)\\
\mathrm{f}^{\mu}\left(\left\{ m_{i\to\mu}\right\} _{i\in\partial\mu},\tau\right) & = & \sum_{\sigma_{\partial\mu\setminus j}}\Theta\left(\tau\left(\sum_{i\in\partial\mu}\xi_{i}^{\mu}\sigma_{i}\right)\right)\prod_{i\in\partial\mu}\left(\frac{1+\sigma_{i}m_{i\to\mu}}{2}\right)
\end{eqnarray}
where $\partial\mu$ represents the set of all input variables involved
in node $\mu$, $\sigma_{\partial\mu}=\left\{ \sigma_{i}\right\} _{i\in\partial\mu}$
the configuration of input variables involved in node $\mu$, $m_{\tau\to\mu}$
the message from the output variable $\tau$ to the node $\mu$ (see
figure~\textbf{\ref{app-fig:BP-perc-node}} for reference). With
these, the messages from factor node $\mu$ to the output variable
node $\tau$ can be expressed as:
\begin{equation}
m_{\mu\to\tau}^{t+1}=\frac{\mathrm{f}^{\mu}\left(\left\{ m_{i\to\mu}^{t}\right\} _{i\in\partial\mu},+1\right)-\mathrm{f}^{\mu}\left(\left\{ m_{i\to\mu}^{t}\right\} _{i\in\partial\mu},-1\right)}{\mathrm{f}^{\mu}\left(\left\{ m_{i\to\mu}^{t}\right\} _{i\in\partial\mu},+1\right)+\mathrm{f}^{\mu}\left(\left\{ m_{i\to\mu}^{t}\right\} _{i\in\partial\mu},-1\right)}\label{app-eq:BPfactor_out}
\end{equation}
while the message from factor node $\mu$ to input variable node j
is: 

\begin{equation}
m_{\mu\to j}^{t+1}=\frac{\mathrm{f}_{j}^{\mu}\left(\left\{ m_{i\to\mu}^{t}\right\} _{i\in\partial\mu\setminus j},m_{\tau\to\mu}^{t},+1\right)-\mathrm{f}_{j}^{\mu}\left(\left\{ m_{i\to\mu}^{t}\right\} _{i\in\partial\mu\setminus j},m_{\tau\to\mu}^{t},-1\right)}{\mathrm{f}_{j}^{\mu}\left(\left\{ m_{i\to\mu}^{t}\right\} _{i\in\partial\mu\setminus j},m_{\tau\to\mu}^{t},+1\right)+\mathrm{f}_{j}^{\mu}\left(\left\{ m_{i\to\mu}^{t}\right\} _{i\in\partial\mu\setminus j},m_{\tau\to\mu}^{t},-1\right)}\label{app-eq:BPfactor_in}
\end{equation}

These functions can be computed exactly in $\mathcal{O}\left(N^{3}\right)$
operations, where $N$ is the size of the input, using either a partial
convolution scheme or discrete Fourier transforms. When $N$ is sufficiently
large, it is also possible to approximate them in $O\left(N\right)$
operations using the central limit theorem, as explained in~\cite{braunstein-zecchina}.
In our tests on the committee machine, due to our choice of the parameters,
we used the approximated fast version on the first layer and the exact
version on the much smaller second layer.

In the fast approximated version, eqs.~(\ref{app-eq:BPfactor_out})
and~(\ref{app-eq:BPfactor_in}) become:
\begin{eqnarray}
m_{\mu\to\tau}^{t+1} & = & \erf\left(\frac{a_{\mu}^{t}}{\sqrt{2b_{\mu}^{t}}}\right)\label{app-eq:BPfactor_out-gauss}\\
m_{\mu\to j}^{t+1} & = & m_{\tau\to\mu}^{t}\frac{\mathrm{g}_{\mu\to j}^{t}\left(+1\right)-\mathrm{g}_{\mu\to j}^{t}\left(-1\right)}{2+m_{\tau\to\mu}^{t}\left(\mathrm{g}_{\mu\to j}^{t}\left(+1\right)+\mathrm{g}_{\mu\to j}^{t}\left(-1\right)\right)}\label{app-eq:BPfactor_in-gauss}
\end{eqnarray}

where we have defined the following quantities:
\begin{eqnarray}
a_{\mu}^{t} & = & \sum_{i\in\partial\mu}\xi_{i}^{\mu}m_{i\to\mu}^{t}\\
b_{\mu}^{t} & = & \sum_{i\in\partial\mu}\left(1-\left(m_{i\to\mu}^{t}\right)^{2}\right)\\
\mathrm{g}_{\mu\to j}^{t}\left(\sigma\right) & = & \erf\left(\frac{a_{\mu\to j}^{t}+\sigma\xi_{j}^{\mu}}{\sqrt{2\left(b_{\mu\to j}^{t}\right)}}\right)\\
a_{\mu\to j}^{t} & = & a_{\mu}^{t}-\xi_{j}^{\mu}m_{j\to\mu}^{t}\\
b_{\mu\to j}^{t} & = & b_{\mu}^{t}-\left(1-\left(m_{j\to\mu}^{t}\right)^{2}\right)
\end{eqnarray}

In~\cite{braunstein-zecchina}, eq.~(\ref{app-eq:BPfactor_in-gauss})
was approximated with a more computationally efficient expression
in the limit of large $N$. We found that this approximation leads
to numerical issues with the type of architectures which we used in
our simulation at large values of $\alpha,$ $y$ and $\gamma$. For
the same reason, it is convenient to represent all messages internally
in ``field representation'' as was done in~\cite{braunstein-zecchina},
i.e.~using $h_{\mu\to j}=\tanh^{-1}\left(m_{\mu\to j}\right)$ (and
analogous expressions for all messages); furthermore, some expressions
need to be treated specially to avoid numerical precision loss. For
example, computing $h_{\mu\to\tau}$ according to eq.~(\ref{app-eq:BPfactor_out-gauss})
requires the computation of an expression of the type $\tanh^{-1}\left(\erf\left(x\right)\right)$,
which, when computed naïvely with standard 64-bit IEEE floating point
machine numbers and using standard library functions, rapidly loses
precision at moderate-to-large values of the argument, thus requiring
us to write a custom function to avoid this effect. The same kind
of treatment is necessary throughout the code, particularly when computing
the thermodynamic functions.

The code for our implementation is available at~\cite{codeFBP}.

After convergence, the single-site magnetizations can be computed
as:
\begin{eqnarray}
m_{j} & = & \tanh\left(\sum_{\nu\in\partial j}\tanh^{-1}\left(m_{\nu\to j}\right)+\tanh^{-1}\left(m_{\star\to j}\right)\right)
\end{eqnarray}
and the average overlap between replicas (plotted in Fig.~\ref{fig:BP-symm-break})
as:

\begin{equation}
q=\frac{1}{N}\sum_{j}m_{j}^{2}
\end{equation}

The local entropy is computed from the entropy of the whole replicated
system from the BP messages at their fixed point, as usually done
within the Bethe-Peierls approximation, minus the entropy of the reference
variables. The result is then divided by the number of variables $N$
and of replicas $y$. (This procedure is equivalent to taking the
partial derivative of the free energy expression with respect to $y$.)
Finally, we take a Legendre transform by subtracting the interaction
term $\gamma S$, where $S$ is the estimated overlap between each
replica's weights and the reference:
\begin{equation}
S=\frac{1}{N}\sum_{j}\frac{m_{j\to\star}m_{\star\to j}+\tanh\left(\gamma\right)}{1+m_{j\to\star}m_{\star\to j}\tanh\left(\gamma\right)}
\end{equation}

From $S$, the distance between the replicas and the reference is
simply computed as $\left(1+S\right)/2$.

\subsection{Focusing BP vs Reinforced BP\label{app-sub:fBP-vs-rBP}}

As mentioned in the main text, the equation for the pseudo-self-interaction
of the replicated Belief Propagation algorithm (which we called ``Focusing
BP'', fBP) is (eq.~(\ref{eq:pseudo-reinforcement}) in the main
text):

\begin{equation}
m_{\star\to j}^{t+1}=\tanh\left(\left(y-1\right)\tanh^{-1}\left(m_{j\to\star}^{t}\tanh\gamma\right)\right)\tanh\gamma\label{app-eq:pseudo-reinforcement}
\end{equation}
See also figure~4 in the main text for a graphical description. The
analogous equation for the reinforcement term which has been used
in several previous works is (eq.~(\ref{eq:reinforcement}) in the
main text):
\begin{equation}
m_{\star\to j}^{t+1}=\tanh\left(\rho\tanh^{-1}\left(m_{j}^{t}\right)\right)\label{app-eq:reinforcement}
\end{equation}

The reinforced BP has traditionally been used as follows: the reinforcement
parameter $\rho$ is changed dynamically, starting from $0$ and increasing
it up to $1$ in parallel with an ongoing BP message-passing iteration
scheme. Therefore, in this approach, the BP messages can only converge
(when $\rho=1$) to a completely polarized configuration, i.e.~one
where $m_{j}\in\left\{ -1,+1\right\} $ for all $j$.

The same approach can be applied with the fBP scheme, except that
eq.~(\ref{app-eq:pseudo-reinforcement}) involves two parameters,
$\gamma$ and $y$, rather than one, and both need to diverge in order
to ensure that the marginals $m_{j}$ become completely polarized
as well.

In this scheme, however, it is unclear how to compare directly the
two equations, since in eq.~(\ref{app-eq:pseudo-reinforcement})
the self-reinforcing message $m_{\star\to j}$ is a function of a
cavity marginal $m_{j\to\star}$, while in eq.~(\ref{app-eq:reinforcement})
it is a function of a non-cavity marginal $m_{j}$. In order to understand
the relationship between the two, we take a different approach: we
assume that the parameters involved in the two update schemes ($\gamma$
and $y$ on one side, $\rho$ on the other) are fixed until convergence
of the BP messages. In that case, one can then remove the time index
$t$ from eqs.~(\ref{app-eq:pseudo-reinforcement}),(\ref{app-eq:reinforcement})
and obtain a self-consistent condition between the quantities $m_{\star\to j}$,
$m_{j\to\star}$ and $m_{j}$ at the fixed point:
\begin{equation}
m_{j}=\tanh\left(\tanh^{-1}\left(m_{\star\to j}\right)+\tanh^{-1}\left(m_{j\to\star}\right)\right)
\end{equation}

Therefore eq.~(\ref{app-eq:reinforcement}) in this case becomes
equivalent to:
\begin{equation}
m_{j}=\tanh\left(\frac{1}{1-\rho}\tanh^{-1}\left(m_{j\to\star}\right)\right)\label{app-eq:reinforcement_converged}
\end{equation}
 to be compared with the analogous expression for the fBP case:
\begin{equation}
m_{j}=\tanh\left(\tanh^{-1}\left(m_{j\to\star}\right)+\tanh^{-1}\left(\tanh\left(\left(y-1\right)\tanh^{-1}\left(m_{j\to\star}\tanh\gamma\right)\right)\tanh\gamma\right)\right)\label{app-eq:pseudo-reincorcement-converged}
\end{equation}

\begin{figure}
\includegraphics[width=0.9\columnwidth]{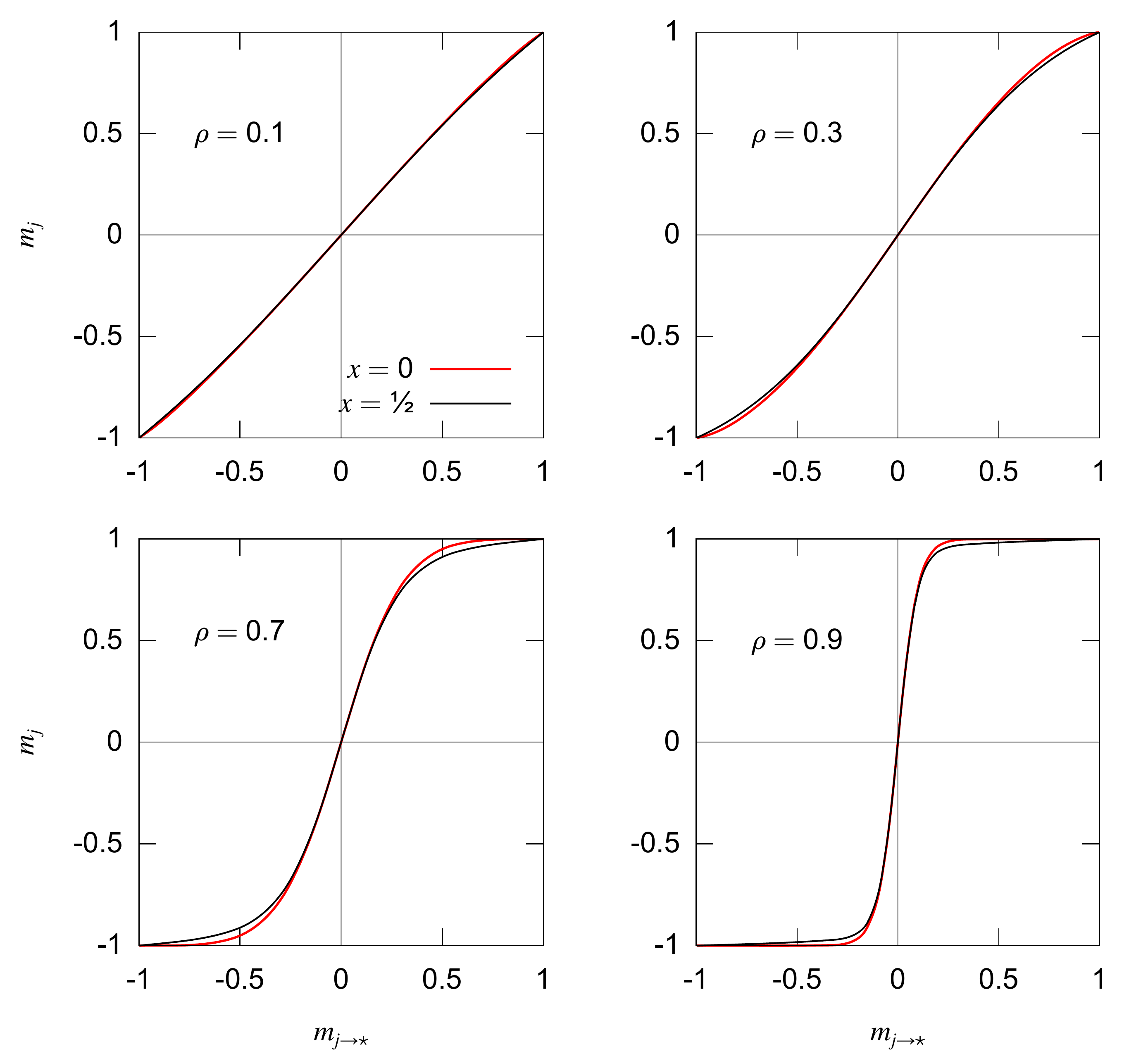}\caption{\label{app-fig:reinf-vs-pseudo}Plots of eq\@.~(\ref{app-eq:pseudo-reincorcement-converged}),
comparison of protocols defined by eqs.~(\ref{app-eq:y-vs-rho})
and~(\ref{app-eq:gamma-vs-rho}) with two different values of the
parameter $x$. The $x=0$ case (thick red lines) corresponds to standard
reinforcement. The curves are in fact very similar across the whole
range of $\rho\in\left[0,1\right]$ and $x\in\left[0,1\right]$, and
consequently display similar performance properties in practice.}

\end{figure}

This latter expression is clearly much more complicated, but by letting
$\gamma\to\infty$ and setting $y=\frac{1}{1-\rho}$ it simplifies
to eq\@.~(\ref{app-eq:reinforcement_converged}). Therefore, we
have an exact mapping between fBP and the reinforced BP. The interpretation
of this mapping in terms of the reweighted entropic measure (eq.~(\ref{eq:part_func})
of the main text) is not straightforward, because of the requirement
$\gamma=\infty$. The $\gamma=\infty$ case at finite $y$ is an extreme
case: as mentioned in the main text, the BP equations applied to the
replicated factor graph (figure~\ref{fig:BPR-scheme} of the main
text) neglect the correlations between the messages $m_{\star\to j}$
targeting different replicas. Since as $\gamma\to\infty$ these messages
become completely correlated, the approximation fails. As demonstrated
by our results, though (figures~\ref{fig:BP-symm-break} and~\ref{fig:fBP-committee}
in the main text, and figure~\ref{app-fig:BPpR-vs-replicas} below),
this failure in the approximation only happens at very large values
of $\gamma$, and its onset is shifted to larger values of $\gamma$
as $y$ is increased, so that even keeping $y$ fixed (but sufficiently
large) and gradually increasing $\gamma$ gives very good results
(figure~\ref{fig:fBP-committee} in the main text). A different approach
to the interpretation of the reinforcement protocol is instead to
consider that it is only one among several possible protocols. As
we showed in the main text for the case of the committee machine,
even keeping $y$ fixed (but sufficiently large) and gradually increasing
$\gamma$ gives very good results. One possible generalization, in
which both $\gamma$ and $y$ start from low values and are progressively
increased, we can consider: 
\begin{eqnarray}
\gamma & = & \tanh^{-1}\left(\rho^{x}\right)\label{app-eq:gamma-vs-rho}\\
y & = & 1+\frac{\rho^{1-2x}}{\left(1-\rho\right)}\label{app-eq:y-vs-rho}
\end{eqnarray}
The second expression was obtained by assuming the first one and matching
the derivative of the curves of eqs.~(\ref{app-eq:reinforcement_converged})
and~(\ref{app-eq:pseudo-reincorcement-converged}) in the point $m_{j\to\star}=0$.
Note that with this choice, both $\gamma\to\infty$ and $y\to\infty$
in the limit $\rho\to1$, thus ensuring that, in that limit, the only
fixed points of the iterative message passing procedure are completely
polarized, and consistently with the notion that we are looking regions
of maximal density ($y\to\infty$) at small distances ($\gamma\to\infty$).
When setting $x=0$, this reproduces the standard reinforcement relations.
However, other values of $x$ produce the same qualitative behavior,
and are quantitatively very similar: figure~(\ref{app-fig:reinf-vs-pseudo})
shows the comparison with the case $x=0.5$. In practice, in our tests
these protocols have proved to be equally effective in finding solutions
of the learning problem.

\subsection{Focusing BP vs analytical results\label{app-subsec:fBP-vs-analytical}}

\begin{figure}

\includegraphics[width=0.9\columnwidth]{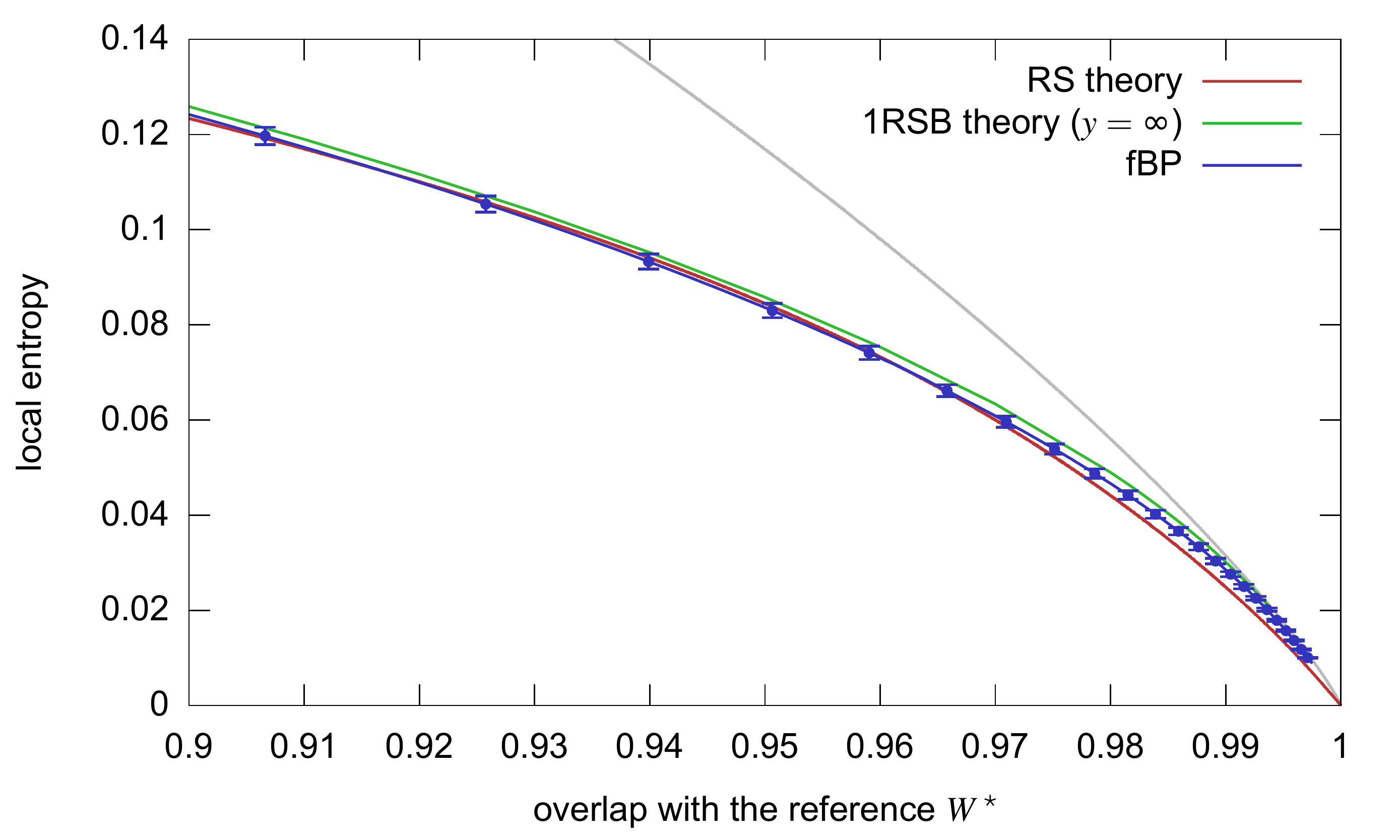}\caption{\label{app-fig:BPpR-vs-replicas}Comparison of local entropy curves
between the fBP results and the analytical predictions, for the case
of the perceptron with $\alpha=0.6$. The algorithmic results (blue
curve) were obtained with $N=1001$ at $y=21$, averaging over $50$
samples. Error bars indicate the estimated standard deviation of the
mean. The RS results (red curve) were also obtained with $y=21$.
The 1RSB results, however, are for the $y=\infty$ case, and it is
therefore to be expected that the corresponding curve is slightly
higher.}

\end{figure}

We compared the local entropy curves produced with the fBP algorithm
on perceptron problems with the RS and 1RSB results obtained analytically
in~\cite{baldassi_subdominant_2015,baldassi2016learning}. We produced
curves at fixed $y$ and $\alpha$, while varying $\gamma$. However,
we only have 1RSB results for the $y=\infty$ case. Figure~\ref{app-fig:BPpR-vs-replicas}
shows the results for $\alpha=0.6$ and $y=21$, demonstrating that
the fBP curve deviates from the RS prediction and is very close to
the 1RSB case. Our tests show that the fBP curve get closer to the
1RSB curve as $y$ grows. This analysis confirms a scenario in which
the fBP algorithm spontaneously choses a high density state, breaking
the symmetry in a way which seems to approximate well the 1RSB description.
Numerical precision issues limited the range of parameters that we
could explore in a reasonable time.

\subsection{Focusing BP on random $K$-SAT\label{app-subsec:fBP-KSAT}}

\begin{figure}
\includegraphics[width=0.9\columnwidth]{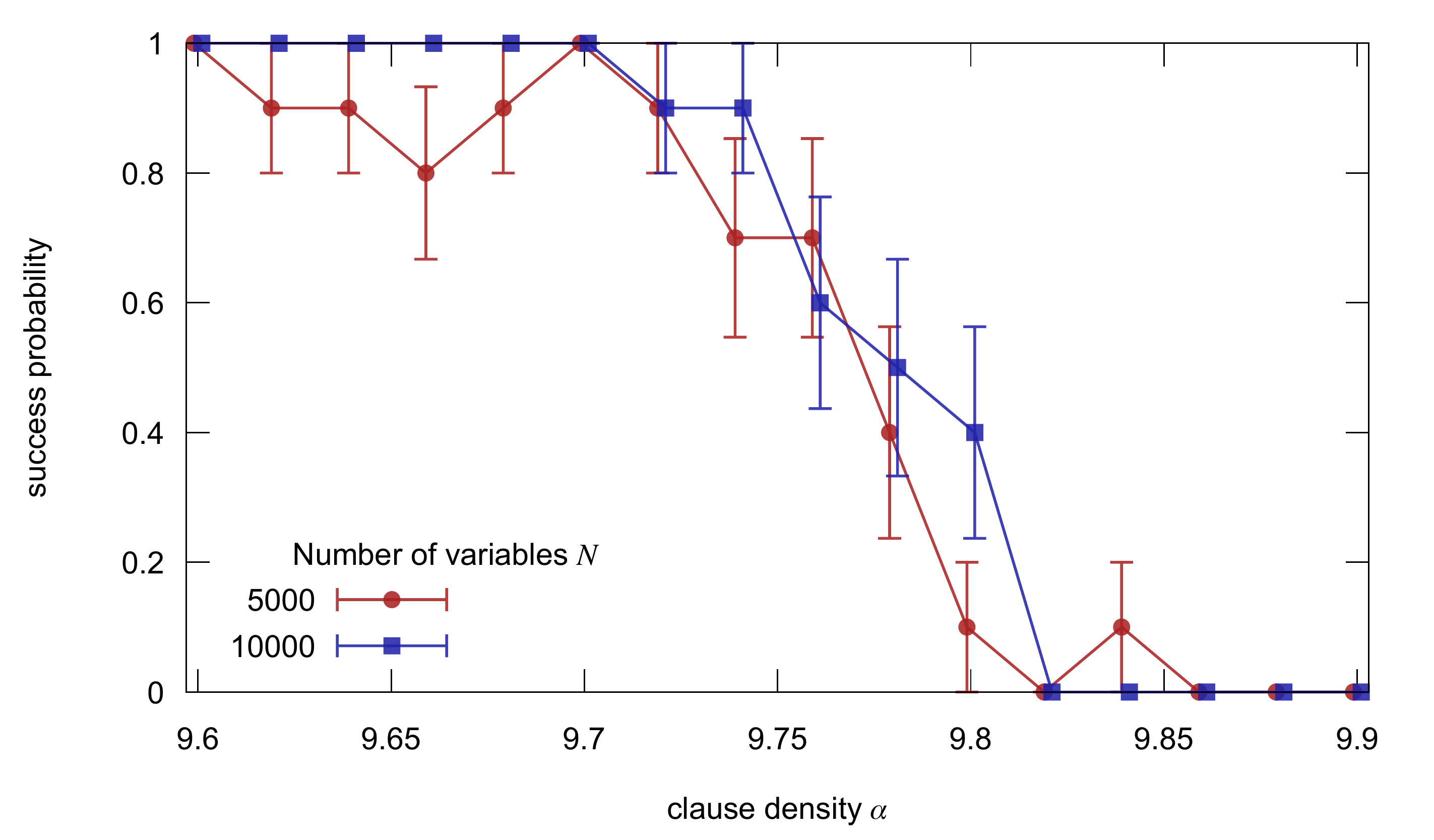}\caption{\label{fig:fBP-ksat}Probability of finding a solution in random instances
of $4$-SAT as a function of clause density $\alpha$ using the fBP
algorithm described in the text. The results are shown for two values
of $N$, slightly shifted relative to each other to improve readability.
The points and error bars represent averages and standard deviations
over $10$ samples.}
\end{figure}

In this section, we show how to apply the Focusing BP (fBP) scheme
on a prototypical constraint statisfaction problem, the random $K$-satisfiability
($K$-SAT) problem \cite{mezard_information_2009}, and present the
results of some preliminary experiments in which it is used as a solver
algorithm.

An instance of $K$-SAT is defined by $N$ variables and $M$ clauses,
each clause $\mu$ involving $K$ variables indexed in $\partial\mu\subseteq\left\{ 1,\dots,N\right\} $,
$\left|\partial\mu\right|=K$. We call $\alpha=M/N$ the clause density.
In the notation of this section, variables are denoted by $\sigma_{i}$,
$i=1,\dots,N$, and take values in $\left\{ -1,+1\right\} $. To each
clause $\mu$ is associated a set of couplings, one for each variable
in $\partial\mu$. The coupling $J_{\mu i}$ takes the value $-1$
if the variable $i$ is negated in clause $\mu$, and $+1$ otherwise.
A configuration $\left\{ \sigma_{i}\right\} _{i=1}^{N}$ satisfies
the clause $\mu$ if $\exists i\in\partial\mu$ such that $J_{\mu i}=\sigma_{i}$.
An instance of the problem is said to be satisfiable if there exists
a configuration which satisfies all clauses. The Hamiltonian for this
problem, which counts the number of violated clauses, is then given
by:

\begin{equation}
H\left(\sigma\right)=\sum_{\mu=1}^{M}\prod_{i\in\partial\mu}\frac{1-J_{\mu i}\sigma_{i}}{2}.
\end{equation}

Random instances of the problem are generated by sampling uniformly
and independently at random the subsets $\partial\mu$ and choosing
the couplings $J_{\mu i}$ independently in $\left\{ -1,+1\right\} $
with equal probability. Random instances of $K$-SAT are locally tree-like
in the large $N$ limit, therefore the problem is suitable to investigation
through the tools of statistical physics of disordered systems, namely
the cavity method \cite{mezard_information_2009}. Such investigation
in fact led in the past decades to outstanding theoretical advances
in understanding the structure of the problem \cite{mezard_information_2009,krzakala-csp}
and to devise state-of-the-art solvers \cite{braunstein2005survey}. 

In order to apply the cavity method to $K$-SAT one identifies each
clause with a factor node and each variable with a variable node;
it is then convenient to parametrize clauses-to-variables cavity messages
with $P_{\mu\to i}\left(\sigma_{i}=J_{\mu i}\right)\equiv\eta_{\mu\to i}$,
and variables-to-clauses ones with $P_{i\to\mu}\left(\sigma_{i}\neq J_{\mu i}\right)\equiv\zeta_{i\to\mu}.$
We also define the variables' subsets $\partial^{+}i$ and $\partial^{-}i$
as $\partial^{\pm}i=\left\{ \mu\in\partial i\ :\ J_{\mu i}=\pm1\right\} $.
With these definitions, the zero temperature BP update rules take
the form:

\begin{align}
\pi_{\pm,i\to\mu}^{t}= & \tilde{\pi}_{\pm,i}^{t}\prod_{\nu\in\partial^{\pm}i\setminus\mu}\,\eta_{\nu\to i}^{t},\label{app-eq:bpksat1}\\
\zeta_{i\to\mu}^{t}= & \frac{\pi_{J_{\mu i},i\to\mu}^{t}}{\pi_{+,i\to\mu}^{t}\ +\pi_{-,i\to\mu}^{t}},\label{app-eq:bpksat2}\\
\eta_{\mu\to i}^{t+1}= & 1-\prod_{j\in\partial\mu\setminus i}\,\zeta_{j\to\mu}^{t}.\label{app-eq:bpksat3}
\end{align}
Magnetizations at time $t$ are given by:

\begin{align}
m_{i}^{t}= & \frac{\tilde{\pi}_{+,i}^{t}\prod_{\nu\in\partial^{+}i}\,\eta_{\nu\to i}^{t}\ -\ \tilde{\pi}_{-,i}^{t}\prod_{\nu\in\partial^{+}i}\,\eta_{\nu\to i}^{t}}{\tilde{\pi}_{+,i}^{t}\prod_{\nu\in\partial^{+}i}\,\eta_{\nu\to i}^{t}\ +\ \tilde{\pi}_{-,i}^{t}\prod_{\nu\in\partial^{+}i}\,\eta_{\nu\to i}^{t}}.\label{app-eq:ksatmag}
\end{align}
The update rules of the coefficients $\tilde{\pi}_{\pm,i}^{t}$ depend
on the algorithm under consideration. In standard BP, the coefficients
$\tilde{\pi}_{\pm,i}^{t}$ are set to:

\begin{equation}
\tilde{\pi}_{\pm,i}^{t}\equiv1.\label{app-eq:pibp}
\end{equation}

A common procedure for finding solutions for the $K$-SAT problem,
called BP guided decimation (BPGD), consists in iterating Eqs.~(\ref{app-eq:bpksat1}-\ref{app-eq:bpksat3})
until convergence; computing marginals according to Eq.~(\ref{app-eq:ksatmag});
fixing a certain fraction of variables, the ones with most polarized
marginals, to $\sigma_{i}=\mathrm{sgn}\left(m_{i}\right)$; repeating
the procedure on the reduced problem, until a satisfying configuration
is found or a contradiction is produced. (A contradiction occurs when
all variables involved in a clause are fixed to values that violate
the clause.)

We can avoid to irrevocably fix the variables at intermediate steps
of the algorithm, as in BPGD, using instead the reinforced BP (rBP)
heuristic, which acts as a ``soft'' decimation. For rBP the new
update rule reads:

\begin{equation}
\tilde{\pi}_{\pm,i}^{t+1}=\left(\pi_{\pm,i}^{t}\right)^{\rho}.\label{app-eq:ksat-reinf}
\end{equation}

In rBP, Eqs.~(\ref{app-eq:bpksat1}-\ref{app-eq:bpksat3}) and (\ref{app-eq:ksat-reinf})
are iterated, while the coefficient $\rho$ is increased up to one.
The algorithm stops when clamping the magnetizations gives a solution,
or a contradiction is produced. 

For fBP, instead, the update rules are derived from the robust ensemble
and are similar to the ones for neural networks:

\begin{align}
m_{i\to\star}^{t} & =\frac{\prod_{\nu\in\partial^{+}i}\,\eta_{\nu\to i}^{t}\ -\ \prod_{\nu\in\partial^{+}i}\,\eta_{\nu\to i}^{t}}{\prod_{\nu\in\partial^{+}i}\,\eta_{\nu\to i}^{t}\ +\ \prod_{\nu\in\partial^{+}i}\,\eta_{\nu\to i}^{t}},\label{app-eq:fbpksat1}\\
m_{\star\to i}^{t+1}= & \tanh\left(\left(y-1\right)\tanh^{-1}\left(m_{i\to\star}^{t}\tanh\gamma\right)\right)\tanh\gamma,\label{app-eq:fbpksat2}\\
\tilde{\pi}_{\pm,i}^{t+1} & =1\pm m_{\star\to i}^{t+1}.\label{app-eq:fbpksat3}
\end{align}

We performed some preliminary tests of the effectiveness of fBP as
a solver in $K$-SAT (see Fig.~\ref{fig:fBP-ksat}), to be expanded
in future investigations. We used the following protocol: we iterated
Eqs.~(\ref{app-eq:bpksat1}-\ref{app-eq:bpksat3},\ref{app-eq:fbpksat1}-\ref{app-eq:fbpksat3})
at fixed $y=6$ and increasing $\gamma$ in steps of $0.01$ every
$2000$ iterations, starting from $\gamma=0.01$. The algorithm is
stopped when a solution is found (by clamping to $\pm1$ the cavity
marginals) or when $\gamma=0.6$ is reached without finding any solution.
The BP update equations often converged to a fixed point before reaching
$2000$ iterations, in which case we skipped to the next value of
$\gamma$; in order to aid convergence, we added some damping to the
iterative procedure.

We didn't test extensively the possible heuristic schemes to turn
fBP into an efficient solver, therefore there is surely room for improvement
over the one adopted here. Nonetheless, our procedure\textemdash when
successful\textemdash found solutions quite rapidly, even though it
is not as fast as rBP. More importantly, it finds solutions up to
high values of $\alpha$. Defining $\alpha_{d}=9.38$, $\alpha_{c}=9.547$
and $\alpha_{s}=9.93$ the dynamic, static and SAT/UNSAT transitions
respectively in $4$-SAT \cite{krzakala-csp}, from Fig.~\ref{fig:fBP-ksat}
it appears that the algorithmic threshold for fBP is $\alpha_{c}<\alpha_{fBP}<\alpha_{s}$.
In comparison, BPGD starts to fail much earlier, $\alpha_{BPGD}\approx\alpha_{d}$
\cite{mezard_information_2009}. The fBP performance seems in fact
to be rather close to that of the ``Survey Propagation with Backtracking''
algorithm of~\cite{marino2015backtrackingSP} (in particular, the
curves in Fig.~1 in that paper can be directly compared and appear
to be similar to those in Fig.~\ref{fig:fBP-ksat} in the present
paper), although a much more extensive and detailed analysis (like
the one performed in~\cite{marino2015backtrackingSP}) would be required
in order to obtain a good estimate of the algorithmic threshold of
fBP in the large $N$ limit.

\end{document}